%% file: main.tex
\documentclass[10pt,twocolumn,letterpaper]{article}

\usepackage{cvpr}              % To produce the CAMERA-READY version
\input{preamble}
\definecolor{cvprblue}{rgb}{0.21,0.49,0.74}
\usepackage[pagebackref,breaklinks,colorlinks,allcolors=cvprblue]{hyperref}

\usepackage{multirow}
\usepackage{graphicx} 
\usepackage{tabularray}
\usepackage{xltabular}
\usepackage{adjustbox} 
\usepackage{amsmath} 
\usepackage{footnotehyper}
\usepackage{threeparttable}
\usepackage{tablefootnote}
\usepackage{booktabs}
\usepackage{tabularx} 
\usepackage{caption} 
\usepackage{siunitx}
\usepackage{diagbox}
\usepackage{makecell} 
\usepackage{pifont}
\usepackage{cuted}
\usepackage[accsupp]{axessibility} 
\usepackage[table]{xcolor} 
\def\paperID{} % *** Enter the Paper ID here
\def\confName{CVPR}
\def\confYear{2026}
\newcolumntype{M}{>{\centering\arraybackslash}m{2.5cm}}
\newcolumntype{Y}{>{\raggedright\arraybackslash}X}

\title{Illuminating Visual Identity in Universal Multimodal Embeddings
}
\newcommand{\linkicon}[1]{\raisebox{-0.15em}{\includegraphics[height=1.05em]{#1}}}
\author{
Jiawei Cao$^{1, 2*}$, 
Junyi Feng$^{2*}$, 
Jiashen Hua$^{2\dagger\ddag}$, 
Ziheng Huang$^{2}$, \\
Bing Deng$^{2}$, 
Kaijie Wu$^{1\ddag}$, 
Chaochen Gu$^{1\ddag}$, 
Jieping Ye$^{2}$ \\
$^{1}$ Shanghai Jiao Tong University $^{2}$ Alibaba Group \\
{\tt\small \{cjw333, kaijiewu, jacygu\}@sjtu.edu.cn, } \\
{\tt\small \{felix.fjy, jiashen.hjs, huangziheng.hzh, dengbing.db, yejieping.ye\}@alibaba-inc.com} \\[6pt]
\href{https://chrisclear3.github.io/MVEB}{\linkicon{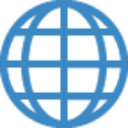}~\texttt{https://chrisclear3.github.io/MVEB}} \\[4pt]
{\small
\href{https://huggingface.co/datasets/hugC/MVEB-train}{\linkicon{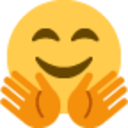}~MVEB-train}\hspace{1.2em}
\href{https://huggingface.co/datasets/hugC/MVEB-test}{\linkicon{images/icons/hf_emoji.png}~MVEB-test}\hspace{1.2em}
\href{https://huggingface.co/HugC/VisME-Qwen25VL-3B}{\linkicon{images/icons/hf_emoji.png}~VisME-3B}\hspace{1.2em}
\href{https://huggingface.co/HugC/VisME-Qwen25VL-7B}{\linkicon{images/icons/hf_emoji.png}~VisME-7B}
}
}

\begin{document}
\maketitle
{
  \renewcommand{\thefootnote}{\fnsymbol{footnote}}
  \footnotetext[1]{Equal contribution. $^\dagger$ Project Leader. $^\ddag$Corresponding author.}
}

\begin{strip}
\centering
\vspace{-17mm} % 缩减间距
\includegraphics[width=\textwidth]{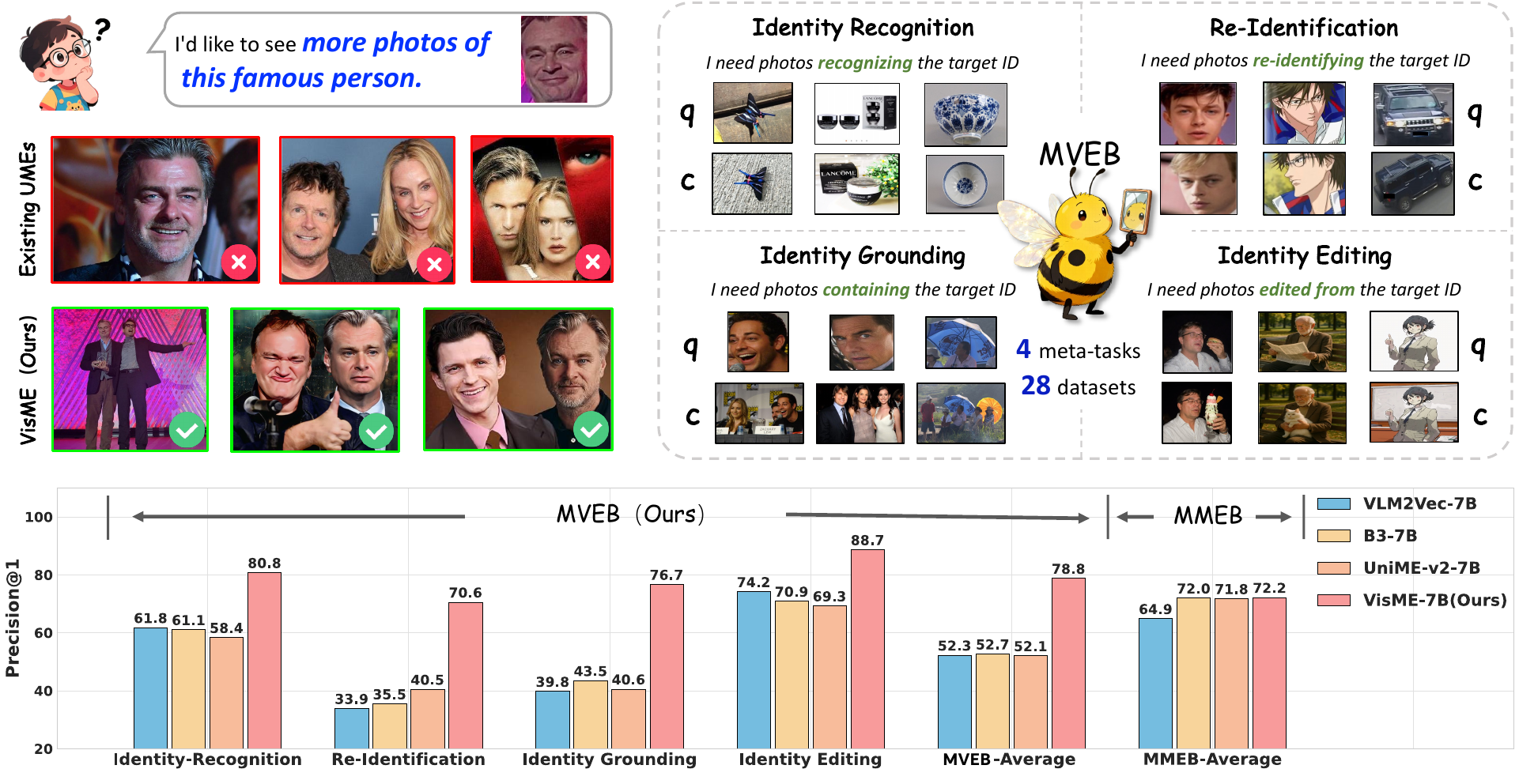}
\vspace{-22pt}
\captionof{figure}{\textbf{Overview of the proposed VisME for identity-level multimodal embeddings.} (a) Qualitative comparison with existing UMEs on visual identity grounding tasks; (b) Illustration of four meta-tasks in MVEB (Identity Recognition, Re-Identification, Identity Grounding, and Identity Editing), covering 28 datasets; (c) Quantitative evaluation on MVEB and MMEB benchmarks.
}
\label{fig:intro}
% \vspace{12mm}
\end{strip}

\input{sec/0_abstract}    
\input{sec/1_intro}
\input{sec/2_related_work}
\input{sec/3_definition}
\input{sec/4_method}
\input{sec/5_experiments}
\input{sec/6_conclusion}

{
    \small
    \bibliographystyle{ieeenat_fullname}
    \bibliography{main}
}
\clearpage 
\appendix
\input{sec/X_suppl}

\end{document}

%% file: sec/0_abstract.tex
\begin{abstract}

Universal Multimodal Embeddings (UMEs) aim to unify various modalities and tasks into a shared representation space. In recent years, this field has witnessed substantial progress driven by the development of Multimodal Large Language Models (MLLMs). 
However, a crucial capability, visual identity discrimination, remains underexplored in existing UME methods, despite its critical role in a wide range of tasks, including instance retrieval, re-identification, and identity preservation in AI-generated content.
To bridge this gap, we propose a unified formulation for visual identity discrimination~(VisID) and introduce $\textbf{MVEB}$ ($\textbf{M}$ultimodal $\textbf{V}$isual Identity $\textbf{E}$mbedding $\textbf{B}$enchmark), a large-scale benchmark curated from both real-world and synthetic datasets to support evaluation and training.
Furthermore, we present a simple yet effective learning framework that jointly optimizes general multimodal and visual identity representations through a carefully designed identity-aware sampling mechanism.
Extensive experiments demonstrate that our approach successfully endows UMEs with strong identity discrimination capability and maintains competitive general multimodal performance.
We believe this work not only illuminates a critical yet neglected capability, but also takes a step toward more holistic universal multimodal embeddings.

\end{abstract}

%% file: sec/1_intro.tex
\section{Introduction}
\label{sec:intro}
Multimodal embedding models aim to encode heterogeneous modalities, such as images, text, and videos, into a unified representation space, enabling practical applications like multimodal search~\cite{cir-fashioniq, magiclens} and multimodal retrieval-augmented generation~\cite{mrag-murag, mrag-visrag}.
% Existing methods generally fall into two paradigms.
% The first paradigm, 
Early approaches, pioneered by CLIP~\cite{CLIP}, adopt a dual-encoder architecture~\cite{align, coca, SigLIP, SigLIP2, llm2clip}, where visual and textual inputs are encoded separately and aligned by contrastive learning on large-scale image–text pairs. 
While effective for cross-modal alignment, these methods typically struggle with complex instructions~\cite{unime} and fused modalities~\cite{magiclens}. 
More recently, universal multimodal embeddings (UMEs)~\cite{vlm2vec, mm-embed, gme, lamra, unime, unite, b3, vlm2vec-v2, jina-v4, qq-embed} powered by multimodal large language models (MLLMs)~\cite{qwen2vl, qwen25vl, llava-ov, phi-3.5-v, internvl3} have been proposed.
By encoding inputs into deep hidden states and performing contrastive fine-tuning on instruction-aware multimodal retrieval datasets~\cite{vlm2vec, mmret-megapairs, mme5}, 
% these approaches inherit key strengths from MLLMs: 
UMEs successfully inherit key capabilities from MLLMs: 
\textit{(i)} a unified embedding space that natively supports fused modalities, 
\textit{(ii)} strong capabilities in long-context reasoning and complex instruction following.
Despite the advances, a fundamental yet overlooked capability remains underexplored in existing UMEs: \textbf{Vis}ual \textbf{I}dentity \textbf{D}iscrimination~(VisID), \ie., the ability to identify images that correspond to the same visual identity as a query image, conditioned on a natural language task instruction
(e.g., “find other photos of this person” or “retrieve images containing the car of the same brand”). 
VisID is a critical capability for both traditional vision-centric representation tasks~(instance retrieval~\cite{uned,udon,ilias,forb}, person re-identification~\cite{reid-survey, reid-scalable, reid-triplet}, face recognition~\cite{arcface, sphereface}) and the recently prominent identity-preserving AI-generated content~\cite{instantid-id-preserving, t2v-id-preserving, magicid-id-preserving, personalvideo-id-preserving}.
However, current SoTA UMEs often fail to distinguish the desired identity, as illustrated in Figure~\ref{fig:intro}.
This is primarily because VisID has not been explicitly addressed in their design.
For example, the most widely used benchmark, MMEB~\cite{vlm2vec}, includes only a single image-to-image subset (NIGHTS~\cite{nights-dreamsim}) among its 36 subsets. 

To bridge this gap, we revisit VisID from the perspective of UMEs and make the following contributions:
\begin{itemize}
    \item 
    \textbf{Problem formulation}. 
    We formally define \textit{Visual Identity Discrimination} as a core capability for UMEs and decompose it into four practical meta-tasks: 
    \textit{(i) Identity Recognition}, 
    \textit{(ii) Re-Identification}, 
    \textit{(iii) Identity Grounding}, 
    and \textit{(iv) Identity Editing}. 
    Some representative cases are shown in Figure~\ref{fig:intro}.
    
    \item 
    \textbf{Benchmark}. 
    We introduce the Multimodal Visual Identity Embedding Benchmark~(MVEB), a large-scale dataset designed to comprehensively evaluate and enhance VisID capabilities. 
    We carefully design a data curation pipeline that integrates both real-world and AI-generated data.
    % Built through a carefully designed data curation pipeline that combines real-world and AI-generated data, 
    As a result, MVEB covers all four VisID meta-tasks across 28 datasets, of which 20 are allocated for training and 8 are reserved for out-of-domain evaluation.
    % , comprising XXX samples in total. 

    \item 
    \textbf{Training framework}. 
    We propose a simple yet effective joint training framework that unifies VisID learning with standard contrastive objectives. 
    Our method employs an identity-aware sampling strategy and a tailored contrastive loss to enforce intra-identity consistency.
    Experiments show that it improves VisID performance by $\sim25\%+$ on MVEB, while maintaining general multimodal retrieval accuracy on MMEB, demonstrating strong compatibility with existing UMEs.

\end{itemize}

We will release data, models, and code to support the development of identity-aware UMEs, a key step toward more capable and general-purpose multimodal representations.

%% file: sec/2_related_work.tex
\section{Related Work}
\label{sec:related_work}

\subsection{Dual-Encoder Multimodal Embeddings}
CLIP~\cite{CLIP} established a foundational paradigm for multimodal representation learning by employing a dual-encoder architecture within a contrastive framework to achieve cross-modal semantic alignment.
Subsequent works refined this paradigm from three main perspectives:
\textit{(i)} Enhancing the text encoder. 
The original CLIP text encoder lacks multilingual capability and is limited to short contexts (maximum 77 tokens).
Chinese-CLIP~\cite{cnclip}, jina-clip-v2~\cite{jina-v2} and SigLIP2~\cite{SigLIP2} introduce multilingual capabilities. 
Recently, with the advent of LLM-based embeddings~\cite{llm2vec, nvembedv2, qwen3embedding}, 
LLM2CLIP~\cite{llm2clip}, SAIL~\cite{sail} and FLAME~\cite{flame} utilize LLM-based text towers~\cite{llm2vec, nvembedv2} that support longer sequences and multiple languages.
\textit{(ii)} Scaling the vision encoder. 
Works such as EVA-CLIP~\cite{evaclip} and InternViT~\cite{internvit} explore larger Vision Transformer (ViT) architectures to improve visual representation capacity.
\textit{(iii)} Improving training objectives. 
BLIP~\cite{BLIP} and CoCa~\cite{coca} incorporate image captioning as an auxiliary training signal, 
while SigLIP~\cite{SigLIP, SigLIP2} advances contrastive learning through optimized training objectives and effective scaling strategies.
Another challenge for the dual-encoder paradigm is the lack of cross-modal fusion ability. 
To tackle composed retrieval tasks,
ALIGN~\cite{align}, UniIR~\cite{uniir} and MagicLens~\cite{magiclens} either simply fuse features by linear addition, or incorporate additional cross-attention modules to realize cross modal interaction.
However, the limited capacity of fusion modules still constrains their performance on modality-compositional tasks.

\subsection{Universal Multimodal Embeddings}
In response to the limitations above, a growing body of research has focused on universal multimodal embeddings~(UMEs)~\cite{e5-v, vlm2vec, mmret-megapairs, mm-embed, gme, mme5, unime, think-then-embed}. 
This emerging paradigm aims to develop a single, unified model capable of processing diverse modalities and generating embeddings for a wide spectrum of tasks. 
From a modeling perspective, most UME approaches fine-tune a multimodal large language model (MLLM)~\cite{llava-ov, qwen2vl, qwen25vl, internvl3} using contrastive learning objectives.
Several notable efforts have been made in training data construction.
M-BEIR~\cite{uniir} aggregates 10 source multimodal datasets to cover 16 diverse retrieval scenarios.
VLM2Vec~\cite{vlm2vec} introduces MMEB, a comprehensive benchmark comprising numerous in-domain and out-of-domain retrieval tasks for systematic evaluation.
MegaPairs~\cite{mmret-megapairs} employs LLMs and VLMs to generate pseudo instructions for curated image pairs, yielding 26M samples.
mmE5~\cite{mme5} curates 560K high-quality multilingual synthetic examples.
Recent works also extend UMEs to more tasks.
GME~\cite{gme} targets visual document retrieval.
UNITE~\cite{unite} and VLM2Vec-v2~\cite{vlm2vec-v2} further incorporate the video modality.
Other approaches focus on performance improvement through mining hard negative samples~\cite{mm-embed, llavae, b3, qq-embed} or introducing specially-designed training stages~\cite{unime, unite}.
While these methods support richer modalities and tasks, they remain focused on semantic-level alignment and largely neglect fine-grained visual identity discrimination. 
In contrast, our approach builds upon the UME paradigm but explicitly prioritizes this critical capability.

\subsection{Visual Identity Discrimination}
Several traditional visual recognition tasks are closely related to visual identity discrimination (VisID), including instance-level image retrieval (e.g., cars~\cite{cars196, compcars}, landmarks~\cite{gldv2, oxford-paris}, e-products~\cite{sop, rp2k}), person re-identification~\cite{market1501, dukemtmc}, and face recognition~\cite{arcface, ms-celeb-1m}.
Classical approaches~\cite{sphereface, reid-scalable, dml} typically learn domain-specific representations.
Although recent works~\cite{unicom, uned, udon, unifex, ilias} aim to build a universal visual embedding model for general-purpose image-to-image search, they remain confined to the \textit{unimodal} setting.
A very recent MLLM-based embedding approach, IDMR~\cite{idmr}, explores grounded instance retrieval.
However, it focuses on a single subtask.
Universal multimodal embeddings still lack a comprehensive formulation and systematic investigation of these identity-centric tasks.
%
% Yet it focuses on a single subtask and lacks a comprehensive formulation or systematic evaluation of VisID as a holistic capability.
% Multimodal embedding frameworks, by contrast, have largely overlooked these identity-centric tasks.
% Multimodal embedding frameworks still 缺少对这些任务的关注.

Visual identity has gained significant attention in AIGC, where ID-preserving image and video generation methods~\cite{instantid-id-preserving, t2v-id-preserving, magicmirror-id-preserving, magicid-id-preserving, personalvideo-id-preserving} aim to retain key attributes of a reference subject during generation.
In contrast to identity preservation in generation, our work focuses on identity discrimination in representations.
We believe that robust identity-aware embeddings can play important roles in AIGC, enabling higher-quality data curation and providing a principled basis for evaluation.

%% file: sec/3_definition.tex
\section{Preliminary: Problem Formulation}
\label{sec:preliminary}
Before introducing our method, we revisit the formulation of universal multimodal embeddings (UMEs) and formally define visual identity discrimination (VisID) within this framework.
Then, we systematically categorize related tasks into four essential meta-tasks that encompass the core aspects of VisID.

UMEs take as input image–text pairs from the multimodal space~$\mathcal{X} = \mathcal{I} \times \mathcal{T}$, where $\mathcal{I}$ and $\mathcal{T}$ denote the spaces of images and text respectively.
The UME model~$f_\theta(\cdot)$ encodes each input sample $x = (i, t)$ into a universal embedding space: $f_{\theta}(x) = y$, where $y\in\mathbb{R}^D$ is a D-dimensional embedding vector.
Within this framework, prior multimodal embedding models~\cite{CLIP,llm2clip} focus on cross-modal alignment between text-only and image-only inputs, i.e., matching $(\epsilon, t)$ with $(i, \epsilon)$, where $\epsilon$ denotes a null~(empty) placeholder indicating the absence of a modality.
In contrast, traditional visual recognition tasks~\cite{uned, reid-survey, arcface} are inherently image-to-image, operating solely on pairs of the form $(i, \epsilon)$.

Building upon the UME framework, we provide a more comprehensive and formal definition of visual identity discrimination (VisID). 
From a capability perspective, VisID refers to the ability to identify images that correspond to the same visual identity as a query image, conditioned on a natural language instruction.
We formalize VisID as a multimodal matching task where both the query and candidate items are represented as $(i, t)$ pairs:

\noindent \textbf{Query $q = (i_q, t_q)$}: The image $i_q$ serves as the identity reference, while the text $t_q$ specifies the target visual content to be identified. The text $t_q$ can be a task instruction (e.g., ``find other photos of this person''), a descriptive caption, or a composition of both.
    
\noindent \textbf{Candidate} $c = (i_c, t_c)$: The image $i_c$ is the primary subject of matching, accompanied by optional text $t_c$ that provides generic context (\eg, ``Represent the image'') or may even be absent ($t_c = \epsilon$).
% \end{itemize}

Based on this formulation, we categorize VisID into four distinct meta-tasks according to the characteristics of the query–candidate pairs.
\begin{itemize}

\item 
\textbf{Identity Recognition.} 
This task requires determining whether two images belong to the same identity, where identity may be defined either as a unique visual instance or as a fine-grained semantic category. It covers tasks such as product recognition, landmark recognition, artifact matching, and fine-grained species recognition.

\item 
\textbf{Re-Identification (Re-ID).} 
Re-ID is a specialized identity recognition task that focuses on matching the same individual entity (e.g., a person, face, or vehicle) across distinct visual observations or depictions.

\item 
\textbf{Identity Grounding.} 
Going beyond conventional visual grounding, this task uses a reference image depicting a known identity as the query, either a cropped instance from a scene or an image of the same entity captured in a different context (\eg, another photo of the same person).

\item \textbf{Identity Editing.} This task addresses the robustness of identity representation against generative transformations. By applying text-prompted edits to a source image, the identity of the subject in the target image is preserved but other attributes (e.g., style, background) are altered.

\end{itemize}
Each of the above meta-tasks presents unique challenges.
Representative examples are shown in Figure~\ref{fig:intro}.

%% file: sec/4_method.tex
\label{sec:method}

\begin{figure*}[t] 
    \centering
    \includegraphics[width=\linewidth]{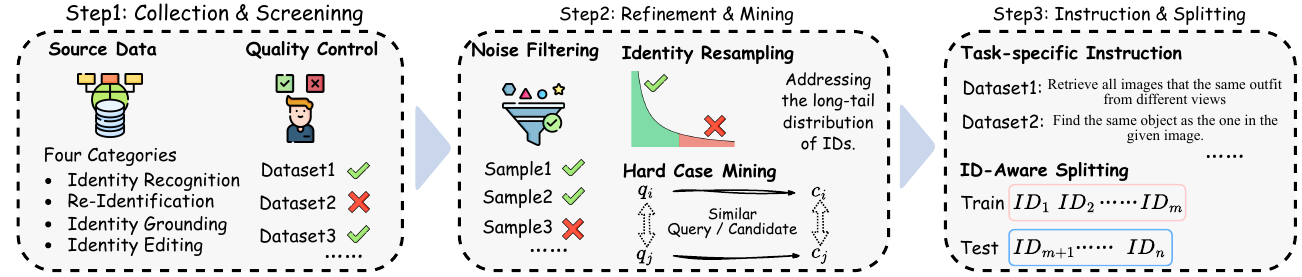}
    \caption{Overall dataset curation pipeline.}
    \label{fig:curation}
\end{figure*}
\begin{figure*}[h] 
    \centering
    \includegraphics[width=\linewidth]{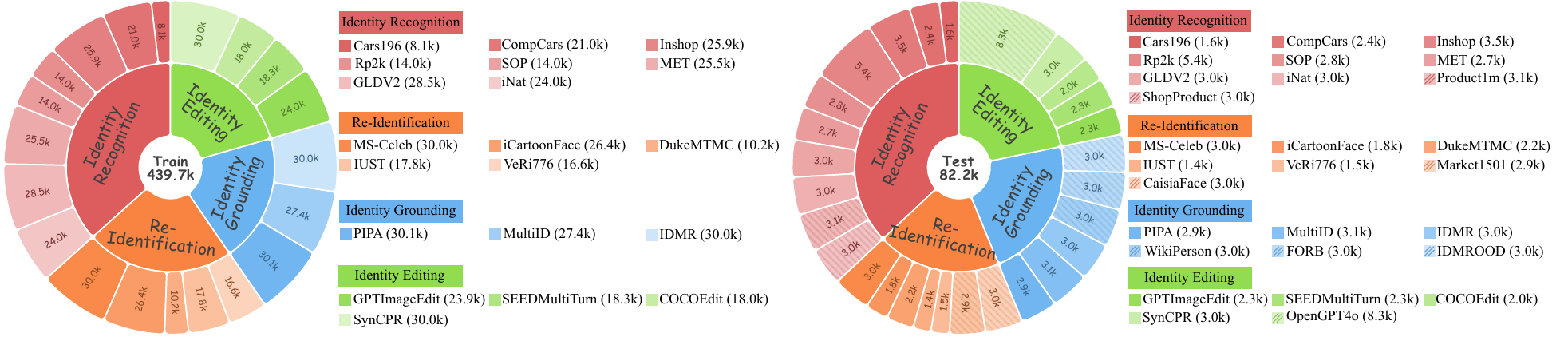}
    \caption{The distribution of MVEB dataset. Sectors shaded in light gray correspond to out-of-domain (OOD) validation sets.}
    \label{fig:data}
\end{figure*}

\section{Methology}

\subsection{MVEB: Data Curation}
To facilitate the training and evaluation of VisID, we construct the \textit{Multimodal Identity Embedding Benchmark} (MVEB), a comprehensive dataset spanning the full spectrum of identity-level tasks defined in Sec.~\ref{sec:preliminary}. 
To ensure both the diversity and the data quality, we carefully design a three-step curation pipeline, illustrated in Figure~\ref{fig:curation}.

\noindent\textbf{Step 1: Collection and Screening.}  
We aggregate raw data from diverse public sources and apply a multi-tier screening process. This includes manual verification of relevance to VisID tasks, followed by dataset-level quality assessment. We exclude entire datasets that fail to meet either the relevance or quality criteria.

\noindent\textbf{Step 2: Refinement and Mining.}  
We perform fine-grained balancing and cleaning in this step:
For existing identity-based datasets, we remove long-tail samples via identity-aware re-sampling to promote domain balance, \eg, for the 4.1M~GLDv2~\cite{gldv2}, we sample 7.5K identities with at most four images each, yielding $\sim$30K training samples.

For AIGC datasets such as GPTImageEdit~\cite{gptimageedit}, where hard-negative samples are scarce, we further leverage an auxiliary embedding model trained following~\cite{unicom,uned,udon} to perform denoising and hard-negative mining.  
% Specifically, the model prunes ambiguous or mislabeled samples during data refinement, and subsequently identifies hard negatives that are semantically aligned with the editing instruction yet originate from different source identities (e.g., generated images of distinct persons wearing the same outfit). This is particularly critical for identity-sensitive tasks like Identity Editing.
Specifically, we first identify candidate negative pairs that share the same editing instruction but originate from different source identities. From this candidate pool, we then employ an auxiliary embedding model to select the pairs whose generated images are closest in the embedding space. These samples are challenging because they are semantically congruent with the instruction yet identity-inconsistent, forcing the model to learn more fine-grained representations.

\noindent\textbf{Step 3: Task Formatting and Splitting.} 
Each sample is assigned a task-specific natural language instruction that explicitly encodes its retrieval objective. (e.g., ``Find the same object as the one in the given image.''). Finally, we partition the data into training (as well as in-domain validation), and out-of-domain validation sets. Crucially, whenever feasible, we enforce a strict \textit{identity-aware split}: no identity present in training appears in any evaluation set. For subdatasets with insufficient unique identities, we fall back to a standard random split at the sample level. 

Through the above curation pipeline, we construct MVEB, a benchmark comprising four meta-tasks, 28 sub-datasets (20 for training and 8 for out-of-domain validation), and a total of 522K samples. Detailed statistics are provided in Figure~\ref{fig:data}.

\subsection{Training framework}
To integrate VisID into existing UME training paradigms, we design a simple yet efficient framework that is fully compatible with standard UME pipelines. Our approach jointly addresses data sampling and loss design, enabling optimization of identity-aware representations in the current framework.

\subsubsection{Data Sampling}
% semantic-aware and identity aware
We consider data sampling from three perspectives, \ie, semantic-aware sampling, identity-aware sampling, and hard-negative sampling to construct meaningful triplets $(q,\mathcal{C}^+,\mathcal{C}^-)$.
Each mini-batch $\mathcal{B}$ is formed by aggregating samples from
$K$ sub-datasets,with each sub-dataset contributing exactly $M$ samples:
\begin{equation}
\mathcal{B} = \bigcup\nolimits_{k=1}^{K} \mathcal{B}_k, \quad |\mathcal{B}_k| = M.
\end{equation}
This design ensures two key properties. 
First, it guarantees \textit{inter-source diversity}, exposing the model to varied data distributions within a single batch. 
Second, it preserves \textit{intra-source coherence}, which naturally promotes the formation of challenging hard negatives. 
Specifically, samples from the same sub-dataset often share high-level semantic or stylistic properties, making them ideal negative candidates.

\noindent \textbf{Semantic-aware Sampling.} For semantic-aware tasks, 
We follow the classic in-batch contrastive learning paradigm. For a given query $q$, the positive set is $\mathcal{C}^+ = \{c^+\}$, where $c^+$ denotes the unique sample in the batch that matches $q$. The negative set comprises all remaining samples in the batch: $\mathcal{C}^- = \mathcal{B} \setminus \{c^+\}$, which includes both intra-source and inter-source negatives.

\noindent \textbf{Identity-aware Sampling.} 
Unlike generic cross-modal datasets, each query $q$ in identity-centric datasets $\mathcal{X}$ belongs to an identity group $\mathcal{G}_q^{\mathcal{X}}$, defined as
\begin{equation}
    \mathcal{G}_q^{\mathcal{X}} = \{ x \in \mathcal{X} \mid \text{ID}(x) = \text{ID}(q) \}.
\end{equation}
Naive in-batch sampling is ill-suited for such data, as it may inadvertently include other members of $\mathcal{G}_q^{\mathcal{X}}$ in the same batch, leading to false-negative pairs that violate the contrastive assumption.

To address this, we adopt an offline sampling scheme: before training, we pre-generate all mini-batches for the entire training schedule, ensuring two key properties: 
(i) each identity appears at most once per batch across all datasets, 
and (ii) the sampling probability of each identity is proportional to the total number of its instances in the dataset.

During training, batches are read sequentially from the pre-scheduled stream. For any anchor $q$, a single positive candidate $c^+$ is then randomly sampled from $\mathcal{G}_q^{\mathcal{X}}$ to form the training triplet.

\noindent \textbf{Structured hard-negative sampling.} 
To enhance the discrimination of models, 
we introduce a mechanism for structured hard-negative mining. 
For certain sub-datasets that contain pre-defined hard-case relationships, we ensure these hard negatives are actively sampled. 
Specifically, when an anchor~$q$ is drawn from such a dataset, we simultaneously sample $k$ of its designated hard negatives into the same mini-batch. 
This forces the model to learn more fine-grained boundaries beyond what random sampling provides.

Consequently, by combining these strategies, both identity-centric and semantic-aware tasks are optimized under a single, consistent objective objective $(q,\mathcal{C}^+, \mathcal{C}^-)$, enabling a unified and effective training framework.

% \subsection{Loss}
\subsubsection{Unified Contrastive Loss}
\label{sec:loss}
To realize our unified learning objective, 
we adopt a single contrastive loss that jointly optimizes semantic alignment and identity discrimination through triplets $(q_i, \mathcal{C}^+_i, \mathcal{C}^-_i)$ generated by our sampling strategy. 
Here, $q_i$ is a query, $c_i^+ \in \mathcal{C}^+_i$ is a positive candidate, and $\mathcal{C}^-_i)$ is the set of negative candidates drawn from the same mini-batch.

Let $f_\theta(\cdot)$ denote the universal embedding function. The similarity between two inputs $x_i$ and $x_j$ as scaled cosine similarity:
\begin{equation}
    Sim(x_i, x_j) = \frac{f_\theta(x_i)^\top f_\theta(x_j)}{\tau}
\end{equation}
where $\tau > 0$ is a learnable temperature parameter.

For each query $q_i$, the loss encourages its embedding to be closer to that of its positive $c_i^+$ than to any negative $c_i^- \in \mathcal{C}^-_i$. Leveraging the natural partitioning of negatives into intra-source and inter-source groups, as ensured by our batch construction, we formulate the loss as:

\begin{equation}
\label{eq:loss}
\mathcal{L}_i = -\log \frac{e^{Sim(q_i, c_i^+)}}{
    e^{Sim(q_i, c_i^+)} 
    + \sum_{c_j^- \in \mathcal{C}^-_i} e^{Sim(q_i, c_j^-)}
}.
\end{equation}

The total batch loss is the average over all queries:
\begin{equation}
    \mathcal{L} = \frac{1}{|\mathcal{B}|} \sum\nolimits_{i \in \mathcal{B}} \mathcal{L}_i.
\end{equation}

This formulation is inherently universal: the same loss (Eq.~\ref{eq:loss}) governs both paradigms, with task specificity emerging solely from how triplets are constructed. In semantic-aware tasks, $c_i^+$ is a caption aligned with $q_i$. 
In identity-discrimination tasks, it is another view or edited variant of the same entity. 
Meanwhile, structured hard negatives ensure fine-grained discrimination, while inter-source negatives provide broad semantic diversity.

%% file: sec/5_experiments.tex
\section{Experiments}
\label{sec:experiments}

\begin{table*}[th]
\centering
\caption{Comparison of methods on benchmark datasets. \textbf{Bold} indicates the best result, \underline{underline} the second best.}
\setlength{\aboverulesep}{0pt}
\setlength{\belowrulesep}{0pt}
\renewcommand{\arraystretch}{1.2}
\label{tab:main_results}
\small
\setlength{\tabcolsep}{5.6pt} 
% \begin{tabular}{@{}l 
%     *{4}{S[table-format=2.1]} 
%     *{4}{S[table-format=2.1]} 
%     S[table-format=2.1] 
%     S[table-format=2.1] 
%     @{}}
\begin{tabular}{@{}l *{10}{c}@{}}
\toprule
Models 
    & \multicolumn{4}{c}{MMEB} 
    & \multicolumn{4}{c}{MVEB} 
    & \multicolumn{2}{c}{Average score} \\
\cmidrule(lr){2-5} \cmidrule(lr){6-9} \cmidrule(l){10-11}
    & {Cls} & {VQA} & {Ret} & {Grd}  
    & {ID-Rec} & {Re-ID} & {ID-Grd} & {ID-Edit}
    & {MMEB$_{\text{avg}}$} & {MVEB$_{\text{avg}}$} \\
\midrule
\multicolumn{11}{c}{\textit{\textbf{Dual-Encoder Models}}} \\
\midrule
CLIP (ViT-large-14)~\cite{vlm2vec}
    & 42.8 & 9.1 & 53.0 & 51.8
    & 55.3 & 44.3 & 37.4 & 52.5 
    & 37.8 & 48.2 \\
SigLIP2 (so400m-14-384)~\cite{SigLIP2}   
    & 55.7 & 9.1 & 43.6 & 62.2
    & 73.1 & 45.6 & 36.8 & 65.4
    & 39.4 & 57.1 \\
LLM2CLIP (EVA-02)~\cite{llm2clip}     
    & 54.1 & 14.5 & 61.3 & 63.5
    & 65.4 & 47.5 & 38.1 & 72.1
    & 46.5 & 56.2 \\
\midrule
\multicolumn{11}{c}{\textit{\textbf{$\sim$\,2B VLM Models}}} \\
\midrule

VLM2Vec (Qwen2-VL-2B)~\cite{vlm2vec}  
    & 58.7 & 49.3 & 65.0 & 72.9
    & 50.1 & 24.1 & 32.5 & 53.3
    & 59.7 & 40.4 \\
GME (Qwen2-VL-2B)~\cite{gme}  
    & 56.9 & 41.2 & 67.8 & 53.4
    & 62.2 & 39.1 & 33.7 & 72.9
    & 55.8 & 52.2 \\
UniME (Phi-3.5-V)~\cite{unime} 
    & 54.8 & 55.9 & 64.5 & 81.8
    & 44.4 & 31.8 & 33.8 & 70.3
    & 64.2 & 43.6 \\
LLaVE (Aquila-VL-2B)~\cite{llavae}
    & 62.1 & 60.2 & 65.2 & \underline{84.9}
    & 29.1 & 26.6 & 29.4 & 72.9
    & 65.2 & 36.4 \\
VLM2Vec-V2.0 (Qwen2-VL-2B)~\cite{vlm2vec} 
    & 62.9 & 56.3 & 69.5 & 77.3
    & 61.8 & 33.9 & 39.8 & 74.2
    & 64.9 & 52.3 \\
B3 (Qwen2-VL-2B)~\cite{b3}  
    & \textbf{67.0} & 61.2 & \textbf{70.9} & 79.9
    & 53.7 & 13.8 & 31.7 & 59.5
    & \underline{68.1} & 40.1 \\
\rowcolor{gray!20}
\textbf{VisME} (Qwen2-VL-2B)  
    & \underline{66.1} & \underline{62.5} & 69.5 & 78.4
    & \underline{74.7} & \underline{54.4} & \underline{62.0} & \underline{87.3}
    & 67.6 & \underline{69.1} \\
\rowcolor{gray!20}
\textbf{VisME} (Qwen2.5-VL-3B)  
    & 63.3 & \textbf{67.4} & \underline{70.5} & \textbf{87.8}
    & \textbf{78.7} & \textbf{67.7} & \textbf{75.1} & \textbf{87.4}
    & \textbf{69.6} & \textbf{76.7} \\
\midrule
\multicolumn{11}{c}{\textit{\textbf{$\sim$\,7B VLM Models}}} \\
\midrule
VLM2Vec (Qwen2-VL-7B)~\cite{vlm2vec}  
    & 62.6 & 57.8 & 69.9 & 81.7
    & 61.8 & 33.9 & 39.8 & 74.2
    & 65.8 & 52.3 \\
GME (Qwen2-VL-7B)~\cite{gme}     
    & 57.7 & 34.7 & 71.2 & 59.3
    & \underline{63.3} & 43.3 & 39.8 & 74.6
    & 56.0 & 55.3 \\
UniME-v2 (LLaVA-OneVision-7B)~\cite{unimev2}     
    & 65.6 & 68.7 & \underline{73.1} & \underline{90.9}
    & 58.4 & 40.5 & 40.6 & 69.3
    & 71.8 & 52.1 \\
LamRA (Qwen2.5-VL-7B)~\cite{lamra}     
    & 51.7 & 34.1 & 66.9 & 56.7
    & 26.9 & 17.2 & 12.9 & 19.9
    & 52.4 & 20.2 \\
LLaVE (Llava-OV-7B)~\cite{llavae}     
    & 65.7 & 65.4 & 70.9 & \textbf{91.9}
    & 57.8 & 42.0 & 44.9 & 77.2
    & 70.3 & 54.5 \\
B3 (Qwen2-VL-7B)~\cite{b3} 
    & \textbf{70.0} & 66.5 & \textbf{74.1} & 84.6
    & 61.1 & 35.5 & 43.5 & 70.9
    & 72.0 & 52.7 \\

\rowcolor{gray!20}
\textbf{VisME} (Qwen2-VL-7B)      
    & \underline{68.9} & \underline{69.1} & 72.8 & 86.1 
    & 78.6 & \underline{60.6} & \underline{68.6} & \textbf{90.2}
    & \underline{72.1} & \underline{74.0} \\
\rowcolor{gray!20}
\textbf{VisME} (Qwen2.5-VL-7B)      
    & 66.4 & \textbf{70.2} & 72.6 & 90.7
    & \textbf{80.8} & \textbf{70.6} & \textbf{76.7} & \underline{88.7}
    & \textbf{72.2} & \textbf{78.8} \\
\bottomrule
\end{tabular}
\vspace{-0.2cm}
\end{table*}

\subsection{Experimental Setup}
\textbf{Datasets and Metrics.} Our training corpus is constructed by combining the MMEB~\cite{vlm2vec} and our proposed MVEB. This unified collection comprises 40 datasets totaling 1.1M pairs. The MMEB portion contributes 20 in-distribution datasets (662K pairs) covering four general tasks: Classification (Cls), VQA, Retrieval (Ret), and Grounding (Grd). The MVEB portion adds another 20 in-distribution datasets (439K pairs) focusing on four identity-centric tasks: Identity Recognition (ID-Rec), Re-Identification (Re-ID), Visual Identity Grounding (ID-Grd), and Identity Editing (ID-Edit). For evaluation, we assess performance on the MMEB (20 in-distribution and 16 out-of-distribution test sets) and the MVEB benchmark (20 in-distribution and 8 out-of-distribution test sets). We report Precision@1 (P@1) as the primary metric for all tasks.

\noindent\textbf{Implementation Details.} Our model, based on Qwen2.5-VL~\cite{qwen25vl} and Qwen2-VL~\cite{qwen2vl}, is trained on 8 NVIDIA A800 GPUs. We use LoRA (rank=16, alpha=32)~\cite{lora} and GradCache~\cite{gradcache} for memory optimization. We process images at a maximum resolution of 200,704 pixels, corresponding to 256 visual tokens, and limit the maximum sequence length at 768. The model is trained for 3,000 steps using a global batch size of 1024 and a learning rate of $1e^{-4}$. 
% We train for 3,000 steps with a global batch size of 1024 and a learning rate of 1e-4. 
The contrastive loss uses a learnable temperature $\tau$, initialized to 0.02, and $k=5$ hard negatives same as the paper~\cite{b3}. 

\subsection{Main Results}
We compare our method against a range of baselines, including dual-encoder models and recent UMEs. The results are summarized in Table~\ref{tab:main_results}.

\begin{itemize}
\item \textit{Dominant Performance Across Benchmarks.} Our method establishes best results on both MMEB and MVEB benchmarks. On the MMEB, our 7B model achieves an average score of 72.2, outperforming the best (B3). The superiority is even more pronounced on our proposed identity-centric MVEB benchmark. Our 2B model alone scores 69.1, significantly surpassing all baselines, and our 7B model extends this lead, reaching an impressive 78.8. This demonstrates our model's robust general multimodal capabilities and its exceptional strength in fine-grained identity understanding.

\item \textit{Analysis of Dual-Encoder Models.} To accommodate the architectural limitations of dual-encoder models, we modified all tasks except Identity Editing into pure image-to-image retrieval tasks by eliminating the text. For Identity Editing, we generate the final representation by averaging the corresponding visual and text embeddings. Under this adapted protocol, SigLIP2 distinguishes itself as the best-performing dual-encoder on MVEB, especially on the Identity Recognition task. This interesting phenomenon suggests that the dual-encoder architecture also holds promise for VisID tasks.
%We hypothesize that this stems from its masked prediction objective, which promotes the learning of fine-grained visual details that are essential for our visual identity-level tasks.

\item \textit{Shortcomings of Existing UMEs.} Among the UMEs, GME-7B demonstrates the best results. We contend that its performance stems from its diverse UMRB dataset, which likely fosters a more optimal feature space distribution. This view is supported by the VLM2Vec-V2.0 model, whose superior performance over other 2B-scale models can be linked to its training on more extensive data. Nevertheless, a clear pattern emerges: all these UME models lag significantly behind our approach on identity-level benchmarks. This performance discrepancy reinforces our central claim that visual identity understanding represents a critical, yet previously neglected, capability in the multimodal representation learning.
\end{itemize}

\subsection{Ablations Analysis}
In the following ablation studies, our default setting is the Qwen2.5-VL-3B model trained on a combined dataset of MMEB and MVEB.

\noindent\textbf{Effectiveness of Sampling Strategies.}
To evaluate the effectiveness of our sampling strategy, we conduct all experiments using LoRA with rank 16 and alpha 32. The results are presented in Table~\ref{tab:sampling}. Our baseline model is trained using interleaved batch sampling without any constraints on the in-distribution dataset; specifically, each interleaved batch has a size of 64, yielding a total batch size of 1024.
By incorporating ID-aware sampling, we observe substantial performance gains: MVEB’s in-distribution accuracy improves by 11.9 points and its out-of-distribution accuracy by 12.8 points compared to the variant without ID-aware sampling. This improvement stems from the fact that, under standard interleaved sampling, each batch is drawn exclusively from a single sub-dataset, which increases the risk of false negatives, particularly when samples sharing identical IDs appear across different batches. Our ID-aware sampling explicitly mitigates this issue by ensuring that such samples are not treated as negative pairs, thereby preventing the severe performance degradation previously observed in MVEB.
Furthermore, by integrating a controlled proportion of hard negative examples into the training pipeline, both MMEB and MVEB consistently outperform random sampling across both IND and OOD settings. Collectively, these two strategies allow for the unified training of both normal and ID-centric data.
\begin{table}[t!]
\centering
\small
\setlength{\tabcolsep}{5pt} 
\caption{Ablation study on our proposed sampling strategy.}
\label{tab:sampling}
\begin{tabular}{
     c c
    S[table-format=2.1]
    S[table-format=2.1]
    S[table-format=2.1]
    S[table-format=2.1]
}
\toprule
\multirow{2}{*}{\makecell{ID-aware \\ Sampling}} &
\multirow{2}{*}{\makecell{Hard Case\\ Mining}} &
\multicolumn{2}{c}{MMEB} &
\multicolumn{2}{c}{MVEB} \\
\cmidrule(lr){3-4} \cmidrule(lr){5-6}
& &
{IND} & {OOD} & {IND} & {OOD} \\
\midrule
 \ding{55} & \ding{55} & 74.2 & 59.4 & 64.3 & 61.2 \\
 \ding{52} & \ding{55} & 74.2 & 59.8 & 76.2 & 74.0 \\
 \ding{52} & \ding{52} & 75.1 & 60.3 & 77.7 & 74.1 \\
\bottomrule
\end{tabular}
\end{table}

\noindent\textbf{Impact of Training Data.}
\begin{table}[t!]
\centering
\small
\setlength{\tabcolsep}{5pt} 
\caption{Ablation study on the composition of training data. Experiments are conducted on the Qwen2-VL-2B.}
\label{tab:framework}
\begin{tabular}{
    l
    S[table-format=2.1]
    S[table-format=2.1]
    S[table-format=2.1]
    S[table-format=2.1]
}
\toprule
\multirow{2}{*}{\diagbox[height=5ex, width=10em]{\scriptsize Data Formula}{\scriptsize Benchmarks}} &
\multicolumn{2}{c}{MMEB} &
\multicolumn{2}{c}{MVEB} \\
\cmidrule(lr){2-3} \cmidrule(lr){4-5}
& {IND} & {OOD} & {IND} & {OOD} \\
\midrule 	
MMEB & 71.4 & 59.0 & 51.7 & 51.2 \\
MVEB & 41.9 & 43.5 & 72.4 & 64.8 \\	
MMEB+MVEB & 72.3 & 59.2 & 76.8 & 72.6 \\
\bottomrule
\end{tabular}
\end{table}
Under the proposed training framework, the formula of training data influences the model's performance preferences. We compare training on the MMEB and MVEB datasets individually versus on their combined mixture. Results are shown in Table~\ref{tab:framework}. 
% Training solely on MMEB yields a baseline performance on MVEB that is 25.1 (IND) and 21.4 (OOD) points lower than mixed training. Conversely, training only on MVEB degrades performance on both MMEB and MVEB. 
Training on a single benchmark only enhances the corresponding capability, failing to generalize to the other.
In contrast, mixed training consistently improves performance on both datasets, with the gains for MVEB exceeding those for MMEB. This indicates that the two datasets complement each other, and that general mixed training particularly benefits MVEB.

\begin{table}[t!]
\centering
\small
\setlength{\tabcolsep}{5pt} 
\caption{Ablation study on interleave batch size.}
\label{tab:interleave}
\begin{tabular}{
    c
    S[table-format=2.1]
    S[table-format=2.1]
    S[table-format=2.1]
    S[table-format=2.1]
    S[table-format=2.1]
    S[table-format=2.1]
}
\toprule
\multirow{2}{*}{\makecell{Interleave \\ Batch size}} &
\multicolumn{3}{c}{MMEB} &
\multicolumn{3}{c}{MVEB} \\
\cmidrule(lr){2-4} \cmidrule(lr){5-7}
& {IND} & {OOD} & {Avg} & {IND} & {OOD} &  {Avg} \\
\midrule
32 & 74.7 & 59.5 & 69.1 & 76.8 & 72.6 & 75.6 \\
64 & 75.1 & 60.3 & 69.6 & 77.7 & 74.1 & 76.7 \\
128 & 75.2 & 60.5 & 69.7 & 78.1 & 73.5 & 76.8\\
\bottomrule
\end{tabular}
\end{table}

\noindent\textbf{Influence of Interleave Batch Size.}
To further investigate the impact of interleave batch size on model performance, we conduct a controlled ablation study under the full sampling strategy described above, varying the interleaved batch size while keeping all other components fixed. Results are reported in Table~\ref{tab:interleave}. As shown in the table, increasing the interleaved batch size from 32 to 64 yields consistent performance gains across both MMEB and MVEB. However, further increasing the size from 64 to 128 leads to only marginal improvements, and even a slight degradation on the MVEB OOD split, suggesting diminishing returns and potential negative effects from excessive intra-batch homogeneity. Based on this analysis, we select an interleaved batch size of 64 as the optimal trade-off between cross-dataset mixing and preservation of ID-aware structure.

\begin{figure*}[t!] 
    \centering
    \includegraphics[width=\linewidth]{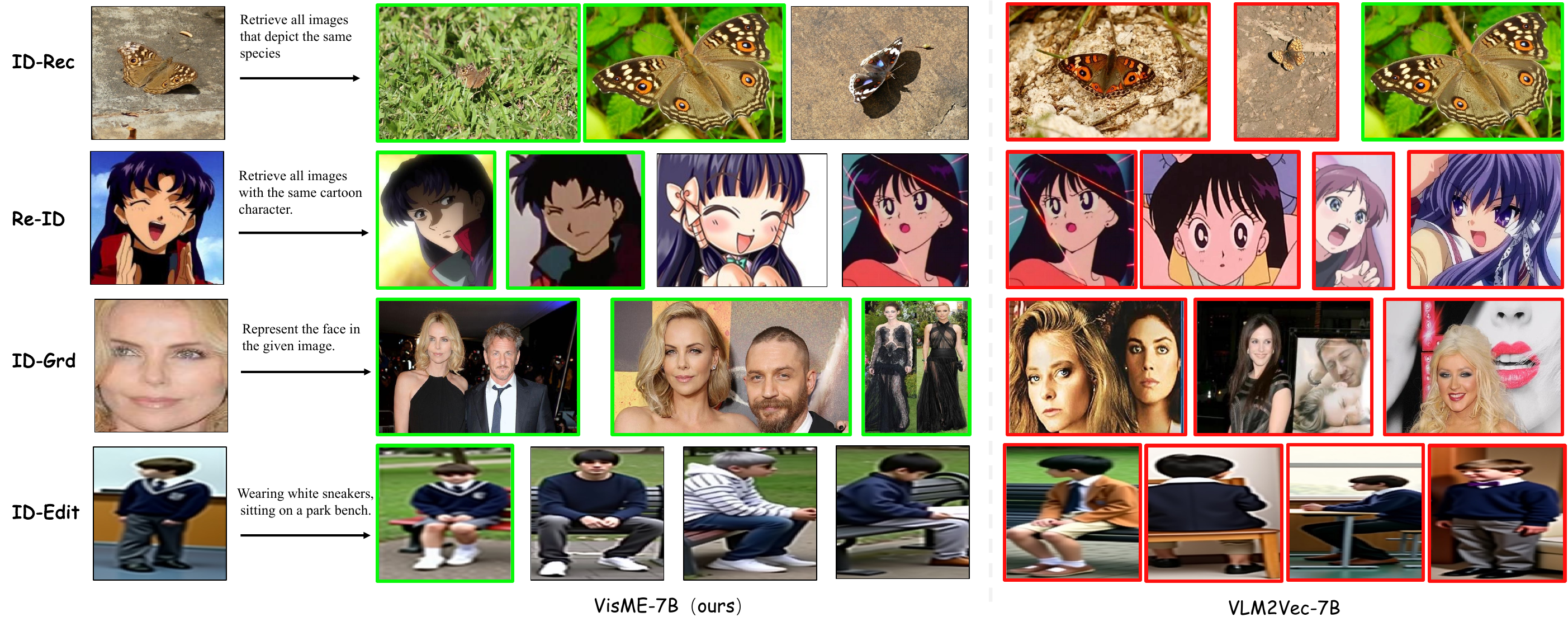}
    \caption{Qualitative comparison of image retrieval results on four tasks (ID-Rec, Re-ID, ID-Grd, and ID-Edit) between our method (based on Qwen2.5-VL-7B) and VLM2Vec-7B. Green boxes denote correct or relevant results, red boxes denote failures or irrelevant outputs, and no boundary indicates that all target images have been retrieved.}
    \label{fig:visual}
\end{figure*}

\begin{table}[t!]
\centering
\small
\setlength{\tabcolsep}{5pt} 
\caption{Ablation study on temperature for contrastive learning.}
\label{tab:temperature}
\begin{tabular}{
    c
    S[table-format=2.1]
    S[table-format=2.1]
    S[table-format=2.1]
    S[table-format=2.1]
    S[table-format=2.1]
    S[table-format=2.1]
}
\toprule
\multirow{2}{*}{\makecell{Temperature \\ $\tau$}} &
\multicolumn{3}{c}{MMEB} &
\multicolumn{3}{c}{MVEB} \\
\cmidrule(lr){2-4} \cmidrule(lr){5-7}
& {IND} & {OOD} & {Avg} & {IND} & {OOD} &  {Avg} \\
\midrule
0.01 & 74.5 & 60.0 & 69.2 & 77.4 & 72.9 & 76.1 \\
0.02 & 75.0 & 60.3 & 69.5 & 77.6 & 74.0 & 76.6 \\
0.03 & 74.5 & 60.0 & 69.1 & 77.5 & 72.0 & 75.9\\
0.02(learnable) & 75.1 & 60.3 & 69.6 & 77.7 & 74.1 & 76.7\\
\bottomrule
\end{tabular}
\end{table}

\noindent\textbf{Sensitivity to Experimental Parameters.} We further conduct ablation studies on key hyperparameters. As shown in Table~\ref{tab:temperature}, initializing the temperature in the loss function at 0.02 yields the best performance, though it is nearly indistinguishable from fixing it at 0.02 throughout training. We also evaluate the effects of batch size and LoRA rank. Increasing the batch size from 256 to 1024 leads to modest performance gains, suggesting that hard-case mining reduces the model’s sensitivity to batch size. Additionally, varying the LoRA rank reveals that rank 32 achieves optimal results.

\begin{figure}
\centering
\includegraphics[width=\linewidth]{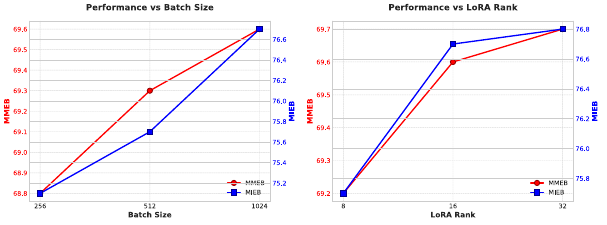}
\caption{Performance variation with respect to batch size and LoRA rank.}
\label{fig:param}

\end{figure}

\subsection{Qualitative Analysis}
Figure~\ref{fig:visual} presents a qualitative analysis of our model’s performance across four meta-tasks, highlighting its superior identity-level capabilities. The model accurately identifies butterfly species by capturing subtle wing patterns—details that baseline models like VLM2Vec overlook. It also exhibits broad cross-domain versatility (Row 2). Notably, our VisME model introduces an identity grounding capability absent in prior UMEs (Row 3): given a cropped face as query, it retrieves both the original source image and other distinct images containing the same person, effectively linking partial observations to a complete identity. Finally, Row 4 shows the model’s proficiency in handling compositional queries involving instruction-based editing, accurately retrieving a specific identity even when the query is modified by textual instructions.

%% file: sec/6_conclusion.tex
\section{Conclusion}
\label{sec:conclusion}
In this work, we addressed a critical yet overlooked capability within the field of Universal Multimodal Embeddings (UMEs): visual identity discrimination (VisID). We argued that the absence of a focus on VisID limits the practical utility of current UMEs. To bridge this gap, we introduced a comprehensive framework encompassing a new problem formulation, a large-scale benchmark targeting this problem (MVEB), and an effective identity-aware training strategy. Extensive experiments show that our approach significantly enhances performance on our identity-centric MVEB by a large margin, while maintaining competitive performance on the general multimodal benchmark, \ie, MMEB. Further, our work may help to establish a quantitative evaluation for identity similarity, boosting future studies about AI-generated content requiring identity preservation. We hope our work will be able to support diverse communities.

\section{Acknowledgments}

This work was supported by the National Natural Science Foundation of China (62227811, 62473255, 62273235). The authors are also affiliated with the Key Laboratory for System Control and Information Processing, Ministry of Education of China and Shanghai Key Laboratory for Perception and Control in Industrial Network Systems.

%% file: sec/X_suppl.tex
\clearpage
\setcounter{page}{1}
\maketitlesupplementary

\section{Model Architecture}
Our VisME utilizes the QwenVL series as the foundational model. The embedding is derived from the hidden states of the last layer corresponding to the last token. During training, we freeze the vision tower and only fine-tune the LLM component with LoRA while maintaining the causal mask. 

\begin{figure}[t]
    \centering
    \includegraphics[width=0.8\linewidth]{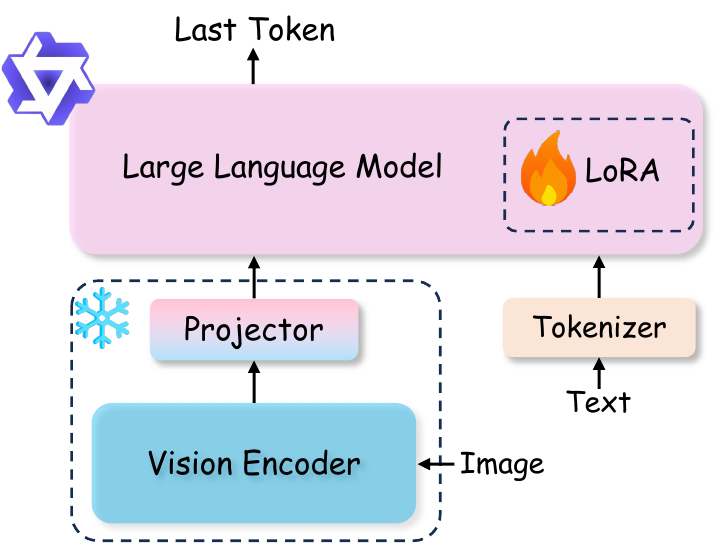}
    \caption{Our VisME model, based on the QwenVL series, freezes the vision components and fine-tunes only the LLM component with LoRA.}
    \label{fig:model}
\end{figure}

\section{Analysis on Identity Discriminability}
\begin{figure}[t]
    \centering
    \includegraphics[width=\linewidth]{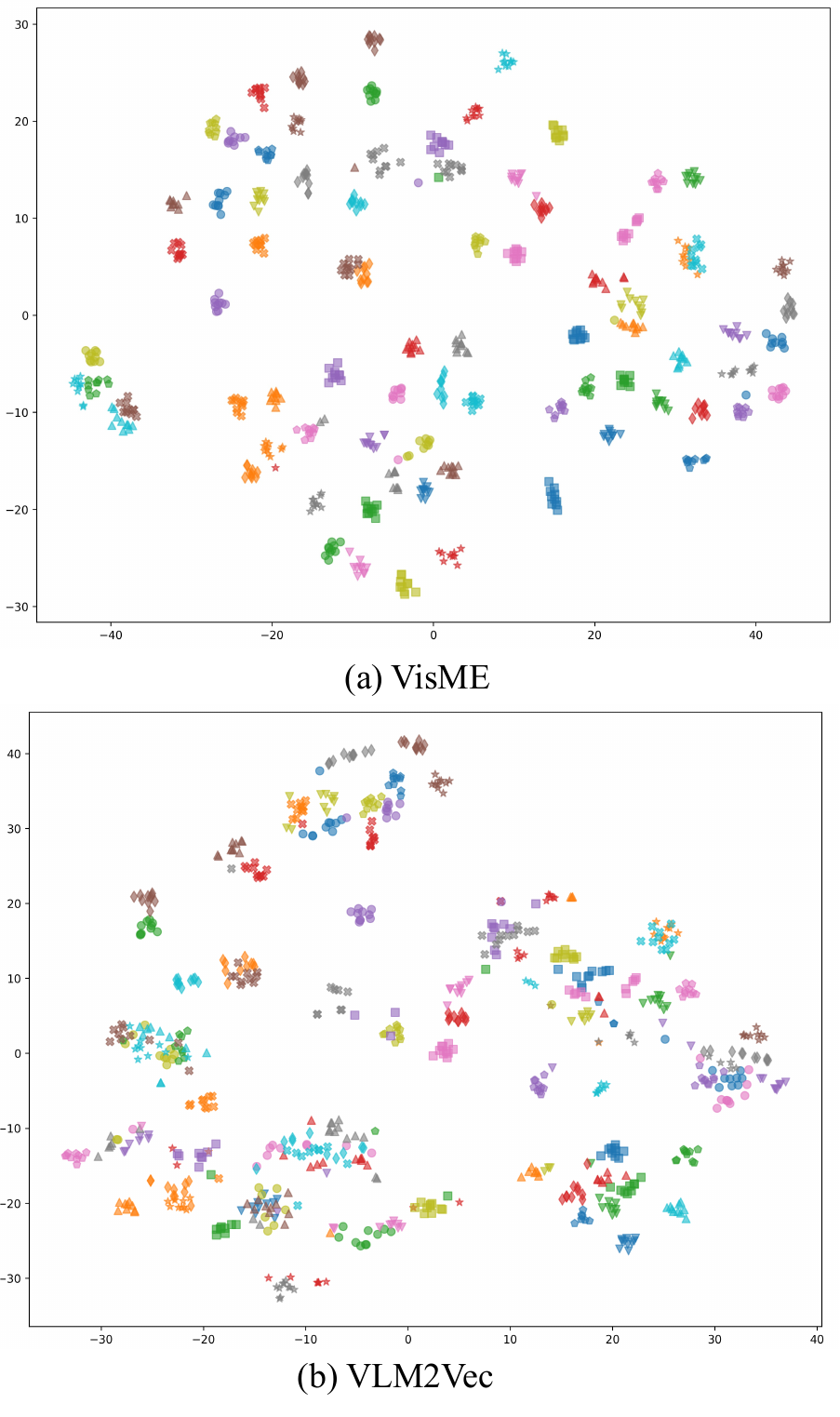}
    \caption{t-SNE of VisME-7B and VLM2Vec-7B embeddings on the extended iNat dataset. Each color–shape combination corresponds to a distinct identity, with 10 samples per identity. }
    \label{fig:tsne}
\end{figure}

To demonstrate identity discriminability of VisME, we visualize the feature distributions of VisME and the baseline model VLM2VEC on a subset of sampled identity (ID) data using t-SNE.
% From the enlarged iNat evaluation set (used in Table~\ref{tab:larger}), 
Specifically, we randomly sample 80 identities~(800 samples) from the enlarged iNat evaluation set (used in Table~\ref{tab:larger}) and employ t-SNE to visualize  feature distributions of VisME-7B and VLM2Vec-7B.  
% we sample 800 samples from 80 identities and employ t-SNE to visualize and compare the embedding distributions of VisME-7B and VLM2Vec-7B, with results shown in Figure~\ref{fig:tsne}.
As shown in Figure~\ref{fig:tsne}, VisME produces significantly more cohesive clusters for each identity, whereas VLM2Vec’s embeddings are notably more dispersed. This provides qualitative evidence that our model effectively discriminates various identities.

\section{Detailed Performance of MVEB}
\begin{table*}[t]
\centering
\caption{Performance of various models on the full MVEB benchmark. Rows shaded in red indicate OOD evaluation sets; rows shaded in gray correspond to meta-task or overall scores.}
\renewcommand{\arraystretch}{1.1}
\label{tab:detail_results}
\small
\setlength{\tabcolsep}{5.6pt} 
\begin{tabular}{@{}l cccccccc@{}} 
\toprule
    Task
    & ID-Unseen?
    & VLM2Vec 
    & GME 
    & LamRA 
    & LLaVE 
    & B3 
    & \makecell{VisME \\ {\scriptsize (Qwen2.5-VL-3B)}} 
    & \makecell{VisME \\ {\scriptsize (Qwen2.5-VL-7B)}}\\
\midrule
\textbf{Identity Recognition}\\
\midrule
Cars196~\cite{cars196}
    & \ding{55} & 63.5 & 72.1 & 36.4 & 55.7
    & 65.7 & 89.2 & 91.8\\
CompCars~\cite{compcars}
   & \ding{52} & 58.1 & 74.2 & 38.7 & 61.1
    & 61.9 & 78.3 & 81.6 \\
Inshop~\cite{inshop}
    & \ding{52} & 61.4 & 75.5 & 12.7 & 80.2
    & 66.8 & 91.6 & 92.2 \\
Rp2k~\cite{rp2k}
    & \ding{52} & 29.3 & 42.2 & 13.8 & 47.3
    & 32.6 & 67.7 & 70.2 \\
SOP~\cite{sop}
    & \ding{52} & 66.1 & 67.0 & 25.6 & 71.9
    & 73.6 & 76.5 & 77.7\\
MET~\cite{met}
    & \ding{52} & 64.5 & 68.2 & 22.3 & 71.0
    & 67.8 & 76.6 & 79.7\\
GLDV2~\cite{gldv2}
    & \ding{52} & 70.8 & 73.3 & 55.8 & 68.1
    & 68.3 & 93.2 & 95.0\\
iNat~\cite{inat}
    & \ding{55} & 29.8 & 40.9 & 30.7 & 33.2
    & 31.9 & 72.2 & 75.8\\
\rowcolor{red!10}
Product1m~\cite{product1m}
    & \ding{52} & 43.0 & 48.1 & 30.2 & 42.0
    & 76.3 & 80.9 & 81.9\\
\rowcolor{red!10}
ShopProduct
    & \ding{52} & 60.3 & 71.6 & 2.7 & 47.6
    & 66.7 & 60.7 & 61.8\\
\rowcolor{gray!20}
All ID-Rec
    & & 61.1 & 63.3 & 26.9 & 57.8
    & 55.3 & 78.7 & 80.8 \\
\midrule
\textbf{Re-Identification}\\
\midrule
MS-Celeb~\cite{ms-celeb-1m}
    & \ding{52} & 21.4 & 29.1 & 18.7 & 24.3
    & 19.2 & 58.8 & 68.5\\
iCartoonFace~\cite{icartoonface}
    & \ding{52} & 29.7 & 49.2 & 6.9 & 58.5
    & 37.0 & 72.5 & 75.8\\
DukeMTMC~\cite{dukemtmc}
    & \ding{52} & 22.3 & 35.7 & 5.4 & 26.7
    & 30.9 & 77.7 & 78.7\\
IUST~\cite{iust}
    & \ding{52} & 33.7 & 35.0 & 12.1 & 33.0
    & 37.6 & 57.8 & 59.4\\
VeRi776~\cite{veri776}
    & \ding{52} & 67.7 & 79.0 & 42.7 & 79.4
    & 68.3 & 82.8 & 83.7\\
\rowcolor{red!10}
Market1501~\cite{market1501}
    & \ding{52} & 34.3 & 57.9 & 22.1 & 56.1
    & 43.5 & 81.0 & 81.3\\
\rowcolor{red!10}
CasiaFace~\cite{casiaface}
    & \ding{52} & 12.1 & 17.2 & 12.6 & 15.9
    & 11.8 & 43.0 & 47.6\\
\rowcolor{gray!20}
All Re-ID
    & & 31.6 & 43.3 & 17.2 & 42.0
    & 35.5 & 67.7 & 70.6 \\
\midrule
\textbf{Identity Grounding}\\
\midrule
PIPA~\cite{pipa}
    & \ding{52} & 17.3 & 12.8 & 1.9 & 27.0
    & 20.4 & 62.4 & 64.6\\
MultiID~\cite{multiid}
    & \ding{52} & 10.0 & 12.5 & 2.9 & 15.4
    & 12.3 & 77.7 & 80.5\\
IDMR~\cite{idmr}
    & \ding{52} & 41.8 & 41.2 & 10.1 & 50.0
    & 51.0 & 70.0 & 71.0\\
\rowcolor{red!10}
WikiPerson~\cite{wikiperson}
    & \ding{52} & 21.4 & 33.6 & 3.8 & 30.5
    & 26.3 & 78.3 & 83.4\\
\rowcolor{red!10}
FORB~\cite{forb}
    & \ding{52} & 81.4 & 84.9 & 35.2 & 82.0
    & 87.2 & 90.8 & 89.7\\
\rowcolor{red!10}
IDMROOD~\cite{idmr}
    & \ding{52} & 58.1 & 53.8 & 23.7 & 64.3
    & 63.5 & 71.5 & 71.3\\
\rowcolor{gray!20}
All ID-Grd
    & & 38.3 & 39.8 & 12.9 & 44.9
    & 43.5 & 75.1 & 76.7 \\
\midrule
\textbf{Identity Editing}\\
\midrule
GPTImageEdit~\cite{gptimageedit}
    & \ding{52} & 77.8 & 84.7 & 27.6 & 91.1
    & 85.5 & 94.6 & 95.0\\
SEEDMultiTurn~\cite{seed}
    & \ding{52} & 53.2 & 73.6 & 19.3 & 63.7
    & 59.8 & 89.6 & 91.1\\
SynCPR~\cite{syncpr}
    & \ding{52} & 49.5 & 62.9 & 5.8 & 74.4
    & 76.0 & 94.1 & 94.0\\
COCOEdit~\cite{uniworld}
    & \ding{52} & 52.2 & 65.7 & 22.0 & 65.8
    & 56.5 & 71.7 & 75.9\\
\rowcolor{red!10}
OpenGPT4o~\cite{opengpt4o}
    & \ding{52} & 71.6 & 86.2 & 24.9 & 90.9
    & 79.8 & 86.8 & 87.5\\
\rowcolor{gray!20}
All ID-Edit
    & & 60.9 & 74.6 & 19.9 & 77.2
    & 70.9 & 87.4 & 88.7 \\
\midrule
\textbf{Final Scores}\\
\midrule
All IND
    & & 46.0 & 54.7 & 20.6 & 54.9
    & 51.0 & 77.7 & 80.1 \\
\rowcolor{red!10}
All OOD
    & & 47.8 & 56.7 & 19.4 & 53.7
    & 56.9 & 74.1 & 75.6 \\
\rowcolor{gray!20}
All 
    & & 46.5 & 55.3 & 20.2 & 54.5
    & 52.7 & 76.7 & 78.8 \\
\bottomrule
\end{tabular}
\end{table*}

This section presents a detailed comparison of VisME against previous UME models, with full results provided in Table~\ref{tab:detail_results}.

The curation pipeline of MVEB guarantees that there is no overlap of specific images between the training and evaluation sets for any sub-dataset, except for the sub-datasets whose original splits contain overlaps. 
This evaluation protocol is specifically designed to assess the model's representation robustness on unseen identities.
% According to the MVEB benchmark's design, only the Cars196 and iNat sub-datasets feature identities that are seen during both training and evaluation. Critically, there is no overlap of specific images between the training and evaluation sets for any sub-dataset. This evaluation protocol is specifically designed to assess the model's representation robustness on unseen identities.

Overall, VisME achieves substantial performance gains over prior UMEs across both IND and OOD settings. The most significant improvements are observed in the Re-Identification and Identity Grounding meta-tasks. This suggests that previous UMEs possess a less explicit capacity for perceiving identity-related information, especially in domains like faces. In contrast, our proposed Visual Identity Grounding task requires models to not only represent a face but also to perceive its specific identity within a global image context. This poses a significant challenge for previous UMEs, as evidenced by their performance on the PIPA, MultiID, and WikiPerson datasets.

Furthermore, scaling our model from 3B to 7B parameters yields consistent improvements. The performance uplift is most pronounced on face recognition datasets, such as MS-Celeb and CasiaFace. This indicates that larger-scale models have a stronger capability for capturing fine-grained identity representations, a crucial attribute for tasks involving human faces.

\begin{table}[t]
    \centering
    \resizebox{\columnwidth}{!}{% 
    \begin{tabular}{l|ccll}
    \toprule
    Sub-Dataset & $\boldsymbol{N_{tgt}}$ & $\boldsymbol{N_{cand}}$ & VLM2Vec & VisME\\
    \midrule
    \multirow{2}{*}{iNat~\cite{inat}} & 2.0 & 3k & 29.8 & 75.8 \\
                          & 9.0 & 100k & 
                          \makecell{17.5 \textcolor{red}{\scriptsize$\downarrow$41.3\%}} & \makecell{56.9 \textcolor{red}{\scriptsize$\downarrow$24.9\%}} \\
    \midrule
    \multirow{2}{*}{SOP~\cite{sop}} & 1.9 & 3k & 66.0 & 77.7 \\
                          & 3.3 & 100k & \makecell{50.6\textcolor{red}{\scriptsize$\downarrow$23.3\%}} & \makecell{67.3\textcolor{red}{\scriptsize$\downarrow$13.4\%}} \\
    \midrule
    \multirow{2}{*}{ShopProduct~\cite{shopproduct}} & 1.0 & 3k & 60.3 & 61.8 \\
                          & 1.0 & 44k & \makecell{46.5\textcolor{red}{\scriptsize$\downarrow$22.9\%}} & \makecell{48.8\textcolor{red}{\scriptsize$\downarrow$21.0\%}} \\
    \midrule
    \multirow{2}{*}{MS-Celeb~\cite{ms-celeb-1m}} & 2.0 & 3k & 21.4 & 68.5 \\
                          & 4.0 & 50k & \makecell{11.4\textcolor{red}{\scriptsize$\downarrow$46.7\%}} & \makecell{57.5\textcolor{red}{\scriptsize$\downarrow$16.1\%}} \\
    \midrule
    \multirow{2}{*}{SynCPR~\cite{syncpr}} & 1.0 & 3k & 49.5 & 94.0 \\
                          & 1.0 & 100k & \makecell{16.9\textcolor{red}{\scriptsize$\downarrow$65.9\%}} & \makecell{67.9\textcolor{red}{\scriptsize$\downarrow$27.8\%}} \\

    \bottomrule
    \end{tabular}
    }
    \caption{Effect of scaling the candidate pool from 3k to up to 100k on retrieval performance for VLM2Vec and VisME across MVEB sub-datasets. Relative performance drops are marked in red.}
    \label{tab:larger}
\end{table}

\section{Analysis on Larger Candidate Pool}
The MVEB benchmark employs a candidate pool of approximately 3k items per sub-dataset, a size selected to balance evaluation diversity with computational feasibility. 
To better mimic real-world deployment settings that involve large-scale candidate galleries, we study the effect of candidate pool size and report performance comparisons on selected sub-datasets with an enlarged candidate set.

As shown in Table~\ref{tab:larger}, expanding the candidate pool leads to a performance decline for both VLM2Vec and VisME. 
Notably, VisME exhibits a smaller relative performance drop (in percentage terms) compared to VLM2Vec. This indicates that VisME not only surpasses existing UME models in absolute performance but also demonstrates superior robustness in large-scale candidate scenarios. 
Nevertheless, the observed performance degradation underscores a key challenge: current models still struggle with retrieval tasks involving large candidate pools. This highlights a substantial direction for future improvement.
To this end, in addition to the standard evaluation configuration used in the main manuscript, we will release a larger-scale version of the MVEB to support the community in conducting more thorough evaluations.

\renewcommand{\tabularxcolumn}[1]{m{#1}}
\begin{table*}[t]
\centering
\caption{Examples of the identity recognition task from MVEB.}
\label{tab:mveb_id_rec}
\begin{tabularx}{\linewidth}{@{}l c c X c@{} r c}
\toprule
\textbf{Dataset} & \textbf{OOD?} & \textbf{Query Image} & \textbf{Query Text} & \textbf{Target Image} & $\quad\boldsymbol{N_{tgt}}$ & $\boldsymbol{N_{cand}}$ \\
\midrule

Cars196~\cite{cars196} & \ding{55}&
\adjustbox{valign=c}{\includegraphics[height=40pt, width=50pt]{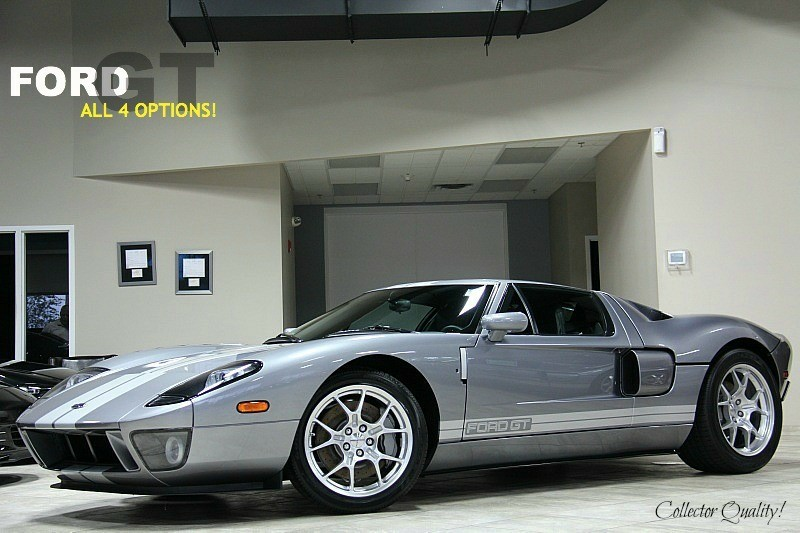}} &
Represent the image with the following text. Find a car of the same model, regardless of color. &
\adjustbox{valign=c}{\includegraphics[height=40pt, width=50pt]{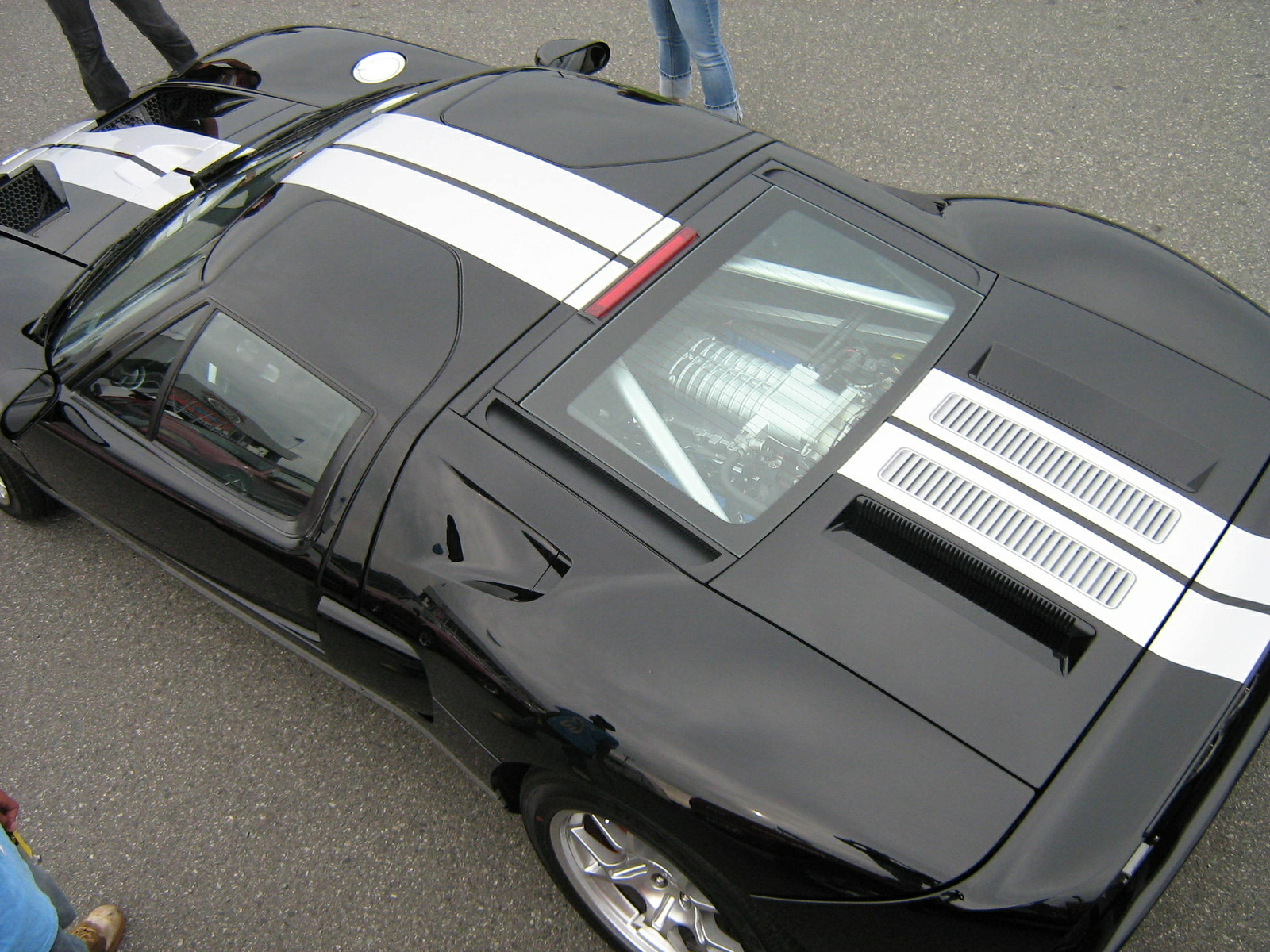}} & 7.0 & 1.5k\\

\midrule
CompCars~\cite{compcars} & \ding{55} &
\adjustbox{valign=c}{\includegraphics[height=40pt, width=50pt]{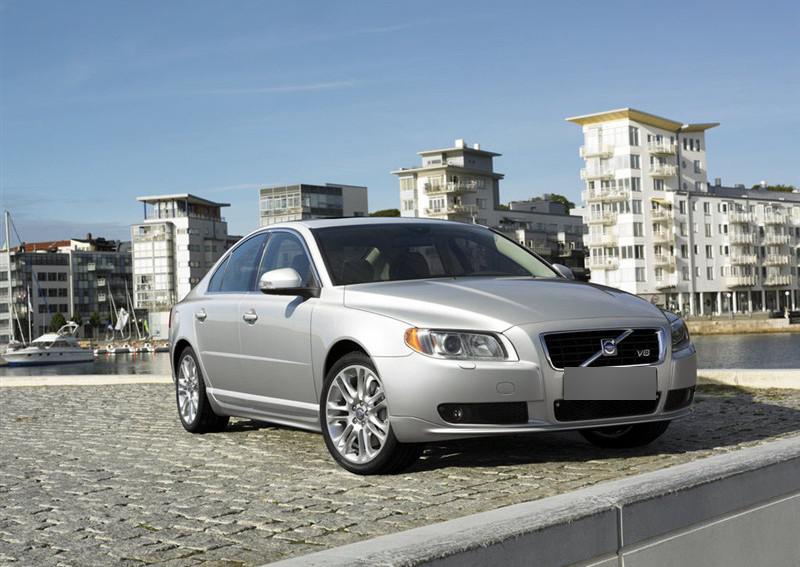}} &
Represent the image with the following text. Find a car of the same model, regardless of color. &
\adjustbox{valign=c}{\includegraphics[height=40pt, width=50pt]{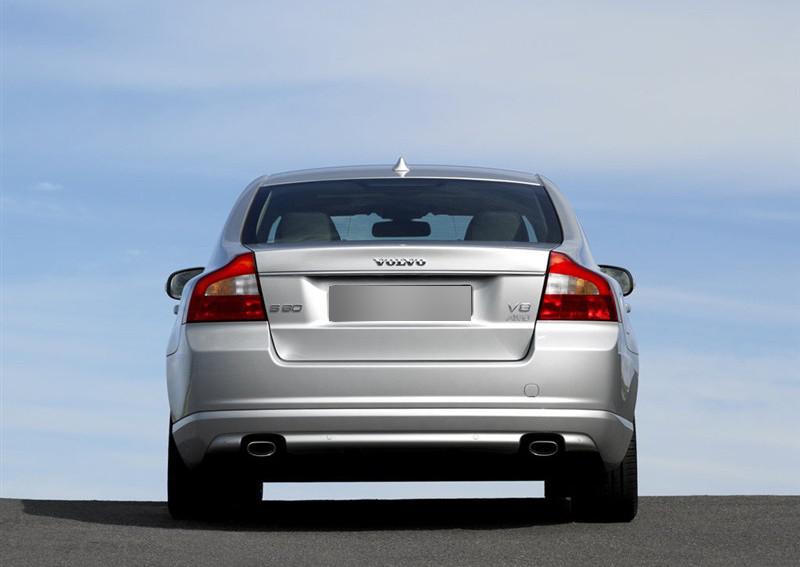}} & 2.0 & 2.3k\\

\midrule
Inshop~\cite{inshop} & \ding{55} &
\adjustbox{valign=c}{\includegraphics[height=40pt, width=50pt]{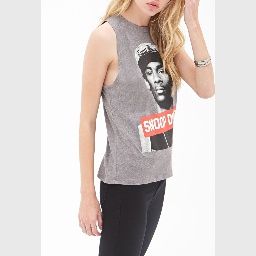}} &
Represent the image with the following text. Retrieve all images that the same outfit from different views. &
\adjustbox{valign=c}{\includegraphics[height=40pt, width=50pt]{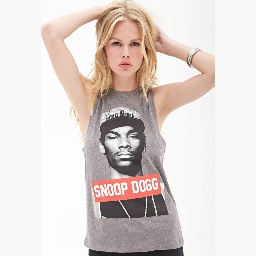}} & 6.2 & 3.1k \\

\midrule
Rp2k~\cite{rp2k} & \ding{55} &
\adjustbox{valign=c}{\includegraphics[height=40pt, width=50pt]{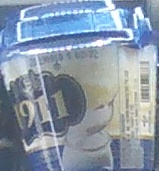}} &
Represent the image with the following text. Find an image of the same product as the one in the given image. &
\adjustbox{valign=c}{\includegraphics[height=40pt, width=50pt]{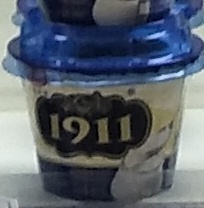}} & 1.9 & 5.4k \\

\midrule
SOP~\cite{sop} & \ding{55} &
\adjustbox{valign=c}{\includegraphics[height=40pt, width=50pt]{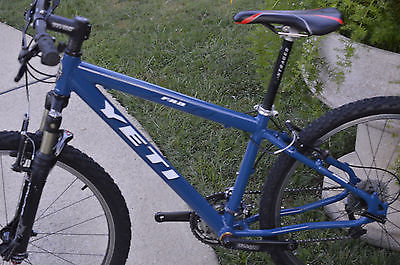}} &
Represent the image with the following text. Find an image of the same product as the one in the given image. &
\adjustbox{valign=c}{\includegraphics[height=40pt, width=50pt]{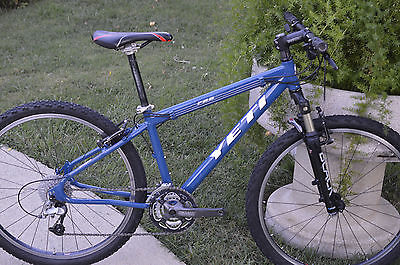}} & 1.9 & 2.8k \\

\midrule
MET~\cite{met} & \ding{55} &
\adjustbox{valign=c}{\includegraphics[height=40pt, width=50pt]{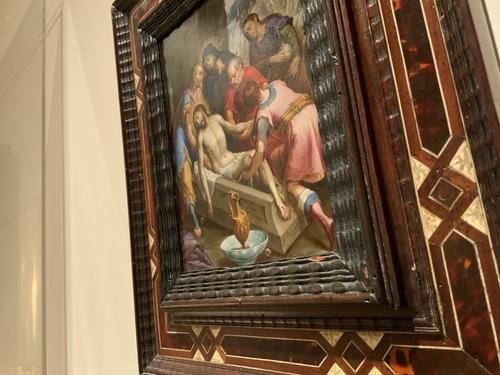}} &
Represent the image with the following text. Find an image of the same artifact as the one in the given image. &
\adjustbox{valign=c}{\includegraphics[height=40pt, width=50pt]{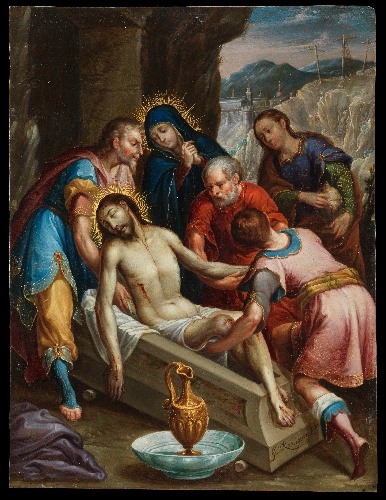}} & 1.8 & 2.7k \\

\midrule
GLDv2~\cite{gldv2} & \ding{55} &
\adjustbox{valign=c}{\includegraphics[height=40pt, width=50pt]{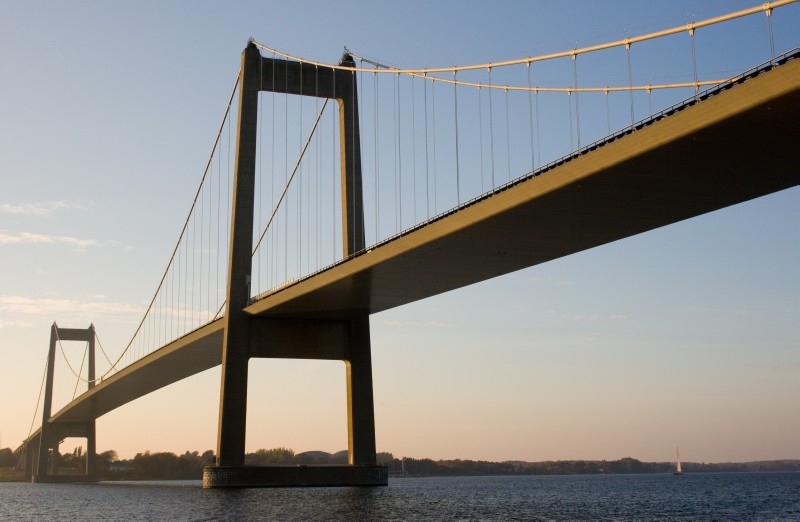}} &
Represent the image with the following text. Retrieve all images that depict the same landmark.&
\adjustbox{valign=c}{\includegraphics[height=40pt, width=50pt]{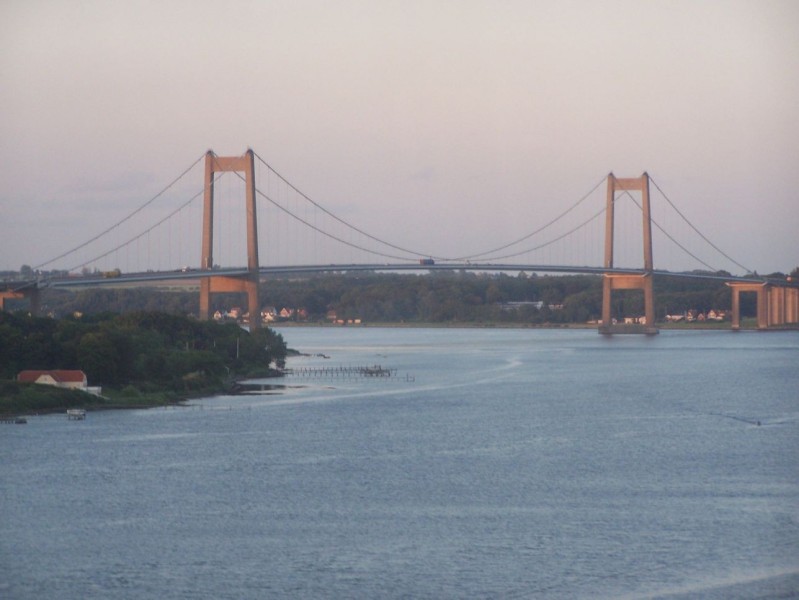}} & 2.0 & 3.0k \\

\midrule
iNat~\cite{inat} & \ding{55} &
\adjustbox{valign=c}{\includegraphics[height=40pt, width=50pt]{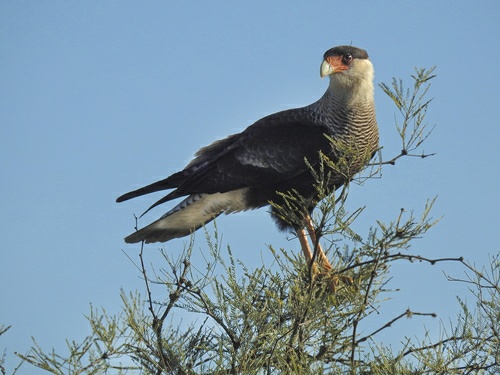}} &
Represent the image with the following text. Retrieve all images that depict the same species.&
\adjustbox{valign=c}{\includegraphics[height=40pt, width=50pt]{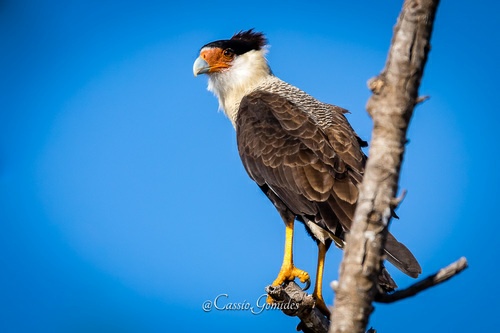}} & 2.0 & 3.0k \\

\midrule
Product1m~\cite{product1m} & \ding{52} &
\adjustbox{valign=c}{\includegraphics[height=40pt, width=50pt]{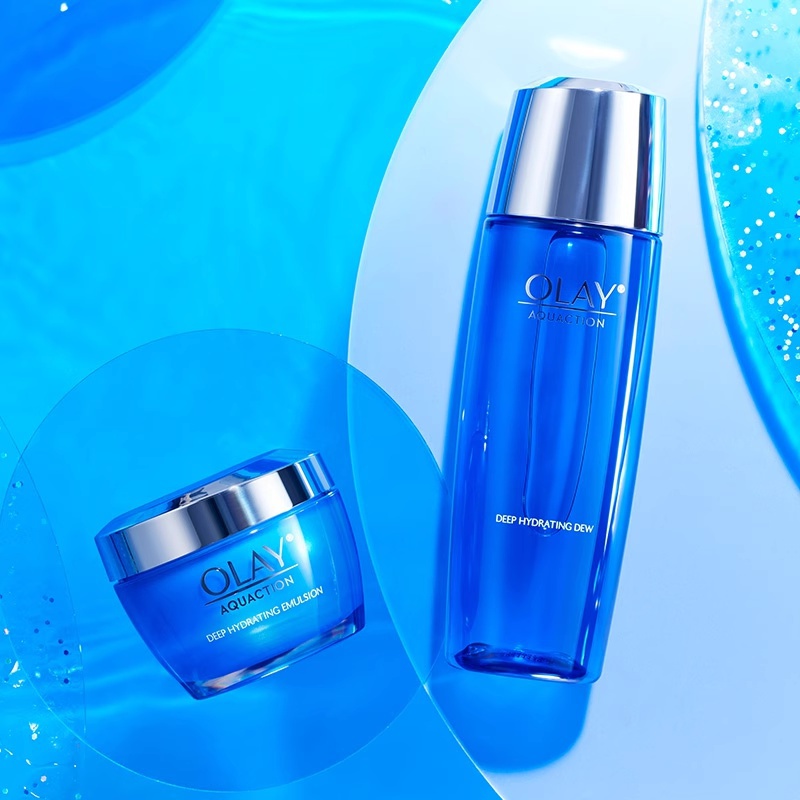}} &
Represent the product in the given image.&
\adjustbox{valign=c}{\includegraphics[height=40pt, width=50pt]{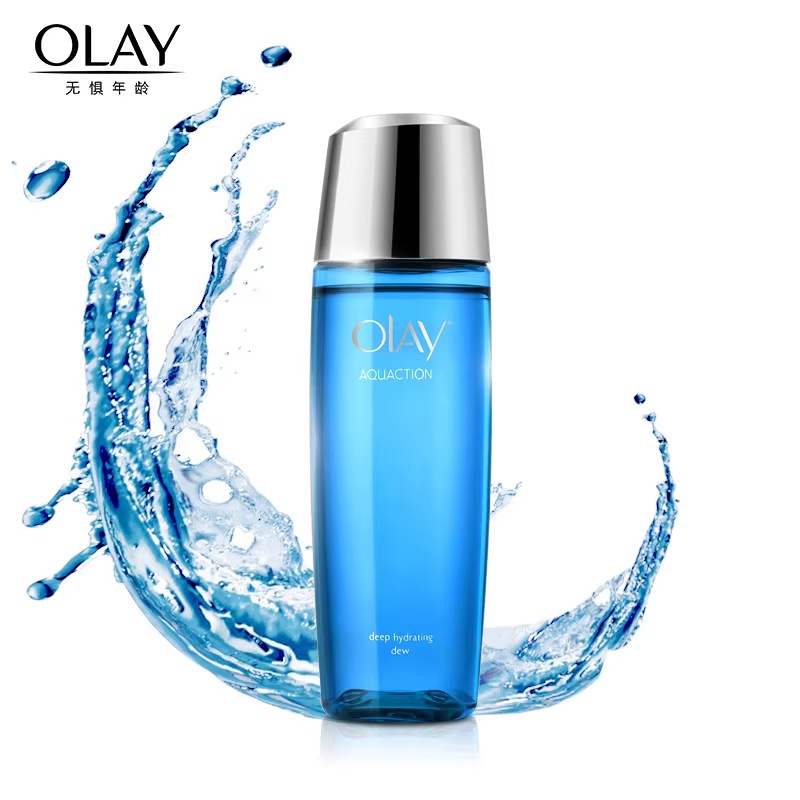}} & 14.2 & 2.2k \\

\midrule
ShopProduct~\cite{shopproduct} & \ding{52}&
\adjustbox{valign=c}{\includegraphics[height=40pt, width=50pt]{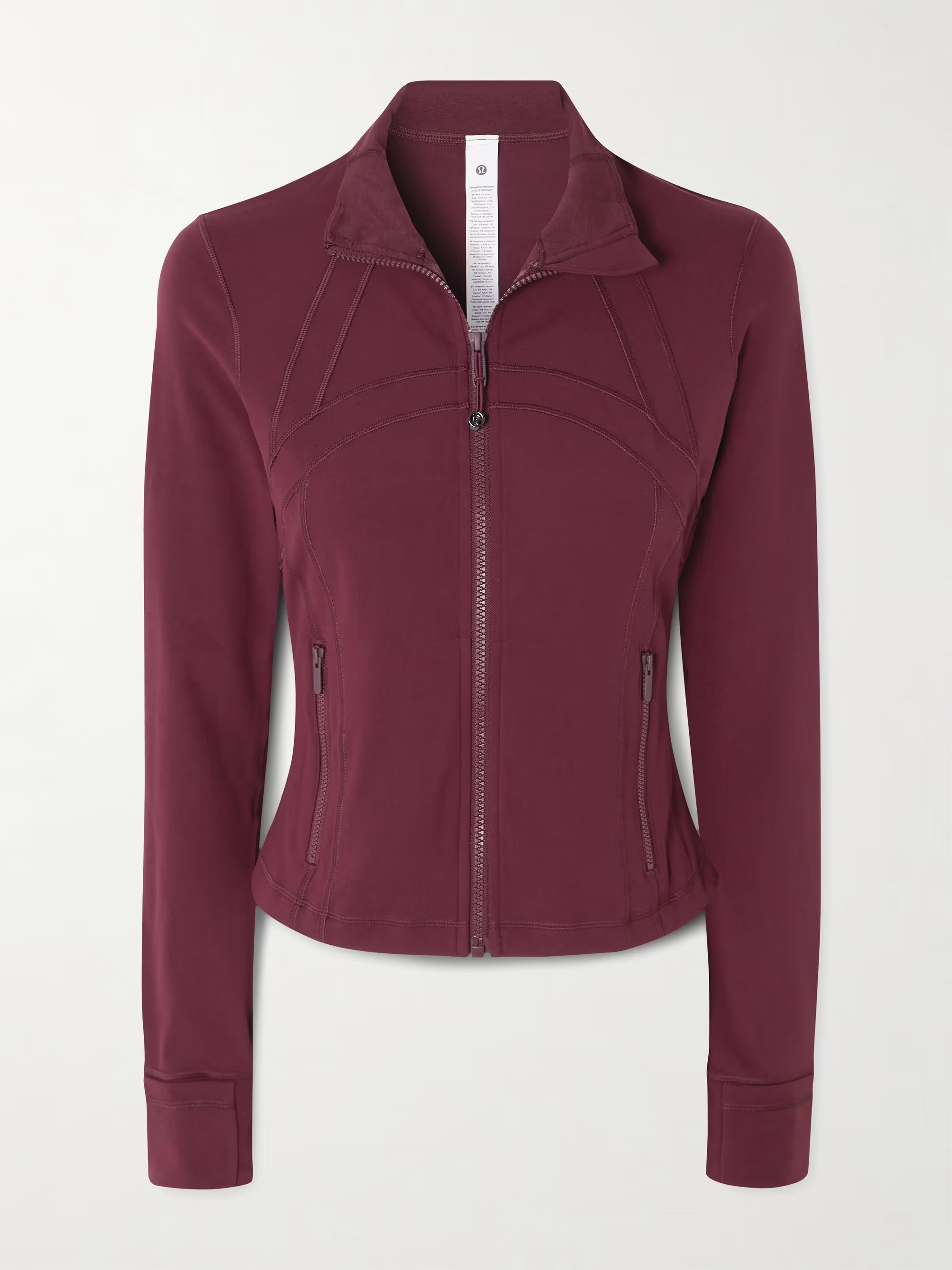}} &
Find an image of a model wearing the product from the given image.&
\adjustbox{valign=c}{\includegraphics[height=40pt, width=50pt]{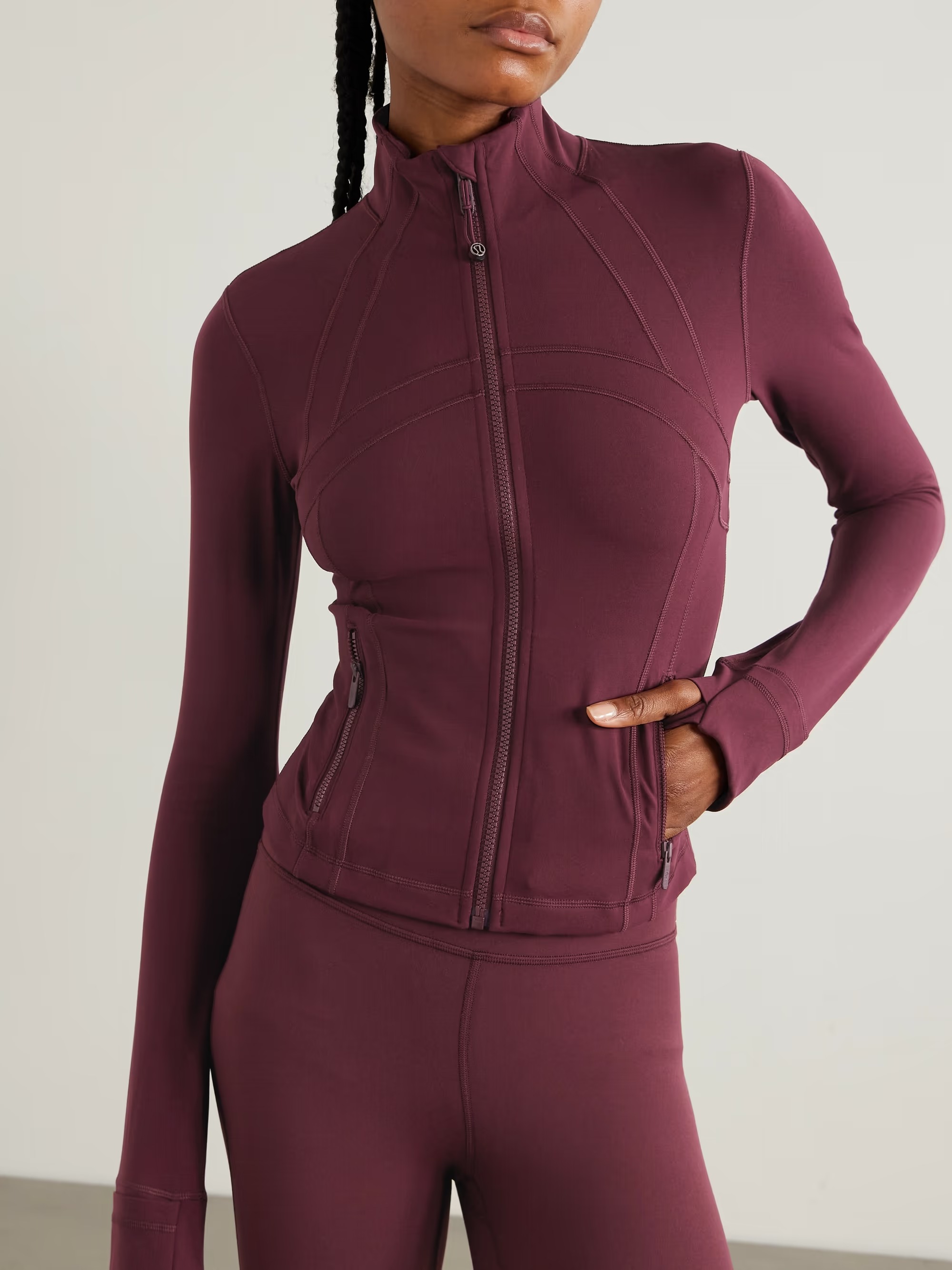}} & 1.0 & 3.0k  \\

\bottomrule 
\end{tabularx}
\end{table*}

\begin{table*}[t] % 由于是table*，[t] 是最有效的选项
\centering
\caption{Examples of the re-identification task from MVEB.}
\label{tab:mveb_reid}
\begin{tabularx}{\linewidth}{@{}l c c X c@{} c c}
\toprule
\textbf{Dataset} & \textbf{OOD?} & \textbf{Query Image} & \textbf{Query Text} & \textbf{Target Image} & $\quad\boldsymbol{N_{tgt}}$ & $\boldsymbol{N_{cand}}$ \\
\midrule
MS-Celeb~\cite{ms-celeb-1m} & \ding{55} &
\adjustbox{valign=c}{\includegraphics[height=40pt, width=40pt]{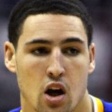}} &
Represent the face with the following text. Retrieve all images with the same facial identity.&
\adjustbox{valign=c}{\includegraphics[height=40pt, width=40pt]{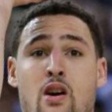}} & 2.0 & 3.0k \\

\midrule
iCartoonFace~\cite{icartoonface} & \ding{55} &
\adjustbox{valign=c}{\includegraphics[height=40pt, width=40pt]{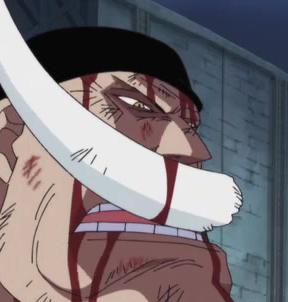}} &
Represent the face with the following text. Retrieve all images with the same cartoon character.&
\adjustbox{valign=c}{\includegraphics[height=40pt, width=40pt]{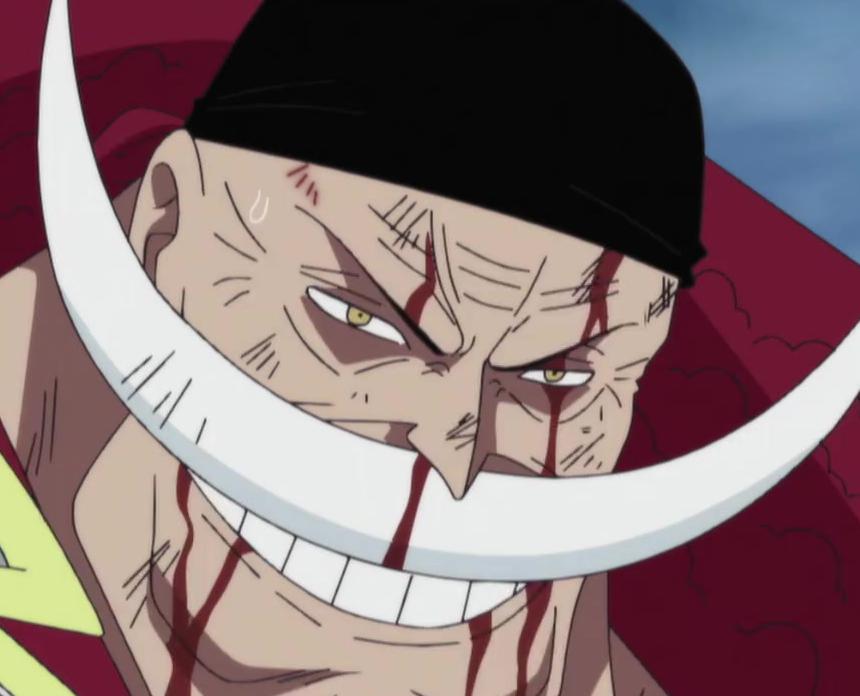}} & 2.0 & 1.8k\\

\midrule
DukeMTMC~\cite{dukemtmc} & \ding{55} &
\adjustbox{valign=c}{\includegraphics[height=40pt, width=40pt]{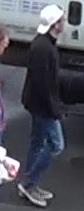}} &
Represent the person with the following text. Re-identify the person in the given image. &
\adjustbox{valign=c}{\includegraphics[height=40pt, width=40pt]{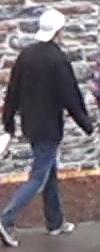}} & 4.0 & 4.4k \\

\midrule
IUST~\cite{iust} & \ding{55} &
\adjustbox{valign=c}{\includegraphics[height=40pt, width=40pt]{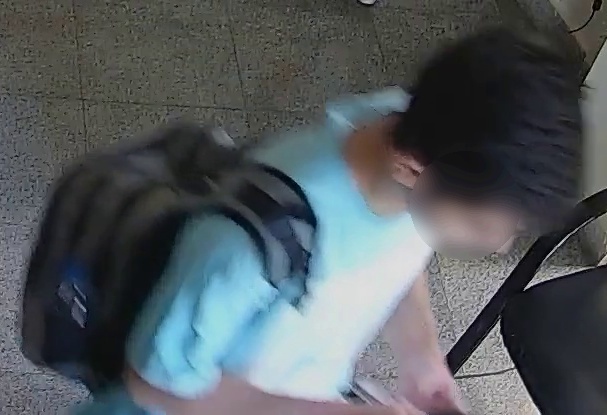}} &
Represent the person with the following text. Re-identify the person in the given image. &
\adjustbox{valign=c}{\includegraphics[height=40pt, width=40pt]{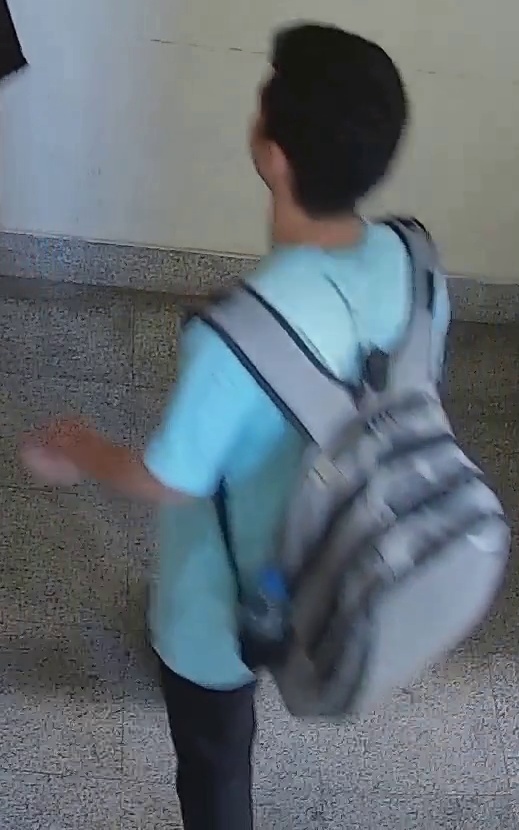}} & 4.0 & 2.6k \\

\midrule
Veri776~\cite{veri776} & \ding{55} &
\adjustbox{valign=c}{\includegraphics[height=40pt, width=40pt]{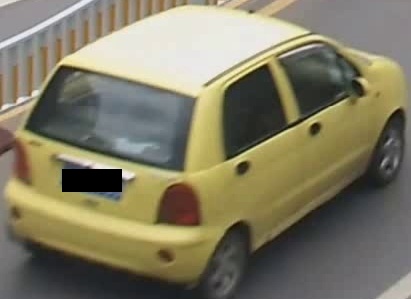}} &
Represent the image with the following text. Retrieve vehicles with the same identity as the query image. &
\adjustbox{valign=c}{\includegraphics[height=40pt, width=40pt]{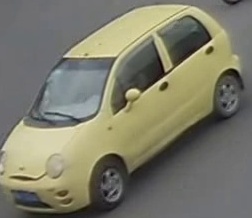}} & 10.0 & 2.6k \\

\midrule
Market1501~\cite{market1501} & \ding{52} &
\adjustbox{valign=c}{\includegraphics[height=40pt, width=40pt]{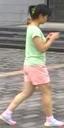}} &
Represent the person with the following text. Re-identify the person in the given image. &
\adjustbox{valign=c}{\includegraphics[height=40pt, width=40pt]{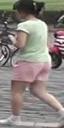}} & 4.0 & 3.0k \\

\midrule
Casiaface~\cite{casiaface} & \ding{52} &
\adjustbox{valign=c}{\includegraphics[height=40pt, width=40pt]{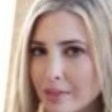}} &
Represent the face with the following text. Retrieve all images with the same facial identity. &
\adjustbox{valign=c}{\includegraphics[height=40pt, width=40pt]{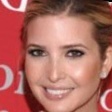}} & 2.0 & 3.0k  \\

\bottomrule 
\end{tabularx}
\end{table*}

\begin{table*}[t] % 由于是table*，[t] 是最有效的选项
\centering
\caption{Examples of the visual identity grounding task from MVEB.}
\label{tab:mveb_id_grd}
\begin{tabularx}{\linewidth}{@{}l c c X c@{} c c}
\toprule
\textbf{Dataset} & \textbf{OOD?} & \textbf{Query Image} & \textbf{Query Text} & \textbf{Target Image} & $\quad\boldsymbol{N_{tgt}}$ & $\boldsymbol{N_{cand}}$\\
\midrule
PIPA~\cite{pipa} & \ding{55} &
\adjustbox{valign=c}{\includegraphics[height=40pt, width=40pt]{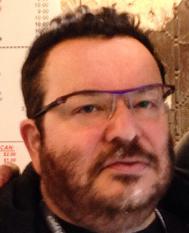}} &
Represent the face in the given image.&
\adjustbox{valign=c}{\includegraphics[height=40pt, width=50pt]{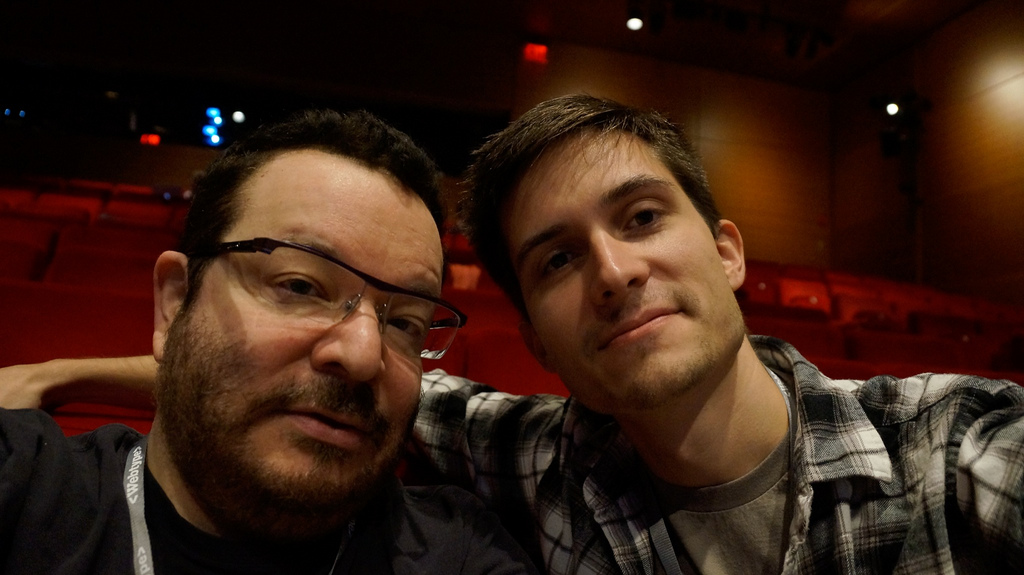}} & 5.0 & 2.9k \\

\midrule
MultiID~\cite{multiid} & \ding{55} &
\adjustbox{valign=c}{\includegraphics[height=40pt, width=40pt]{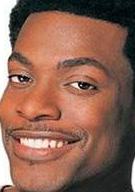}} &
Represent the face in the given image.&
\adjustbox{valign=c}{\includegraphics[height=40pt, width=60pt]{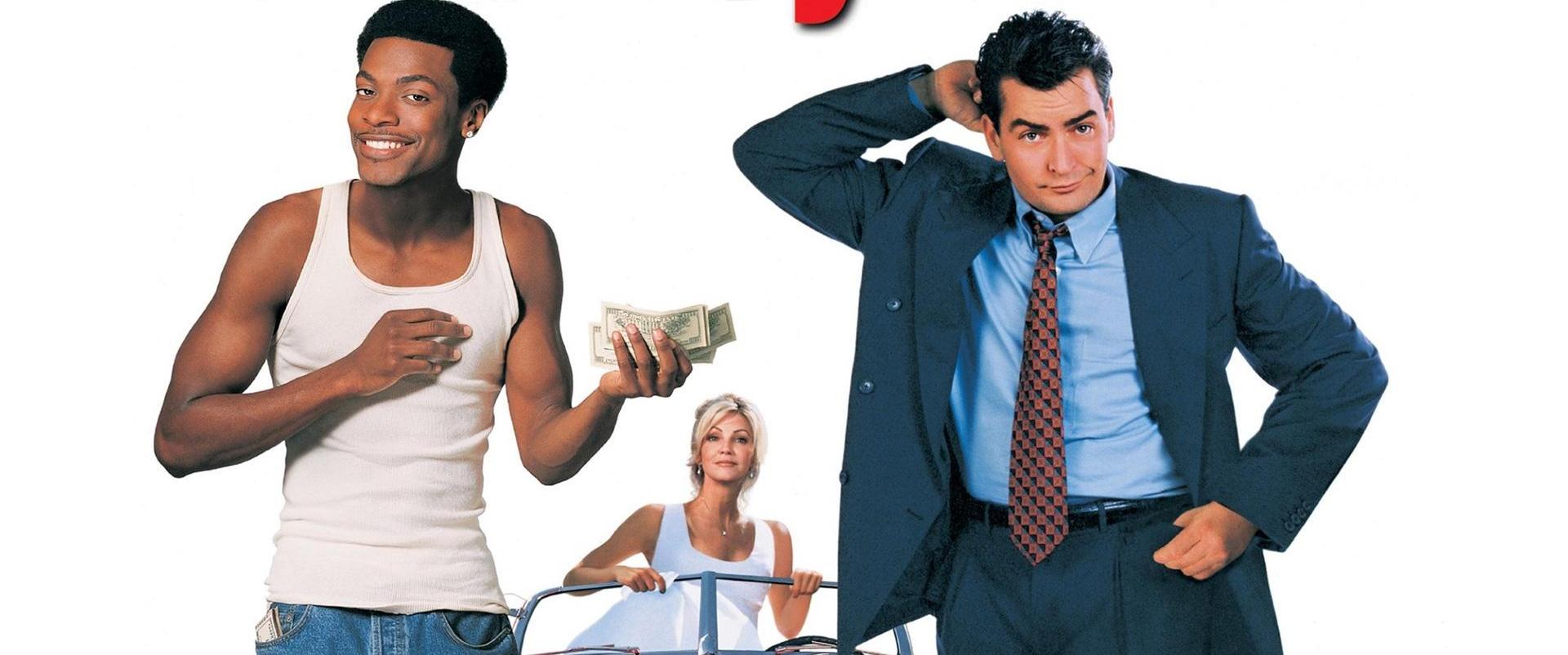}} & 5.8 & 3.1k \\

\midrule
IDMR~\cite{idmr} & \ding{55}&
\adjustbox{valign=c}{\includegraphics[height=40pt, width=40pt]{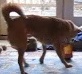}} &
Find me an image containing the object in the given image with the following caption: The dog is standing on a blue rug in front of a television, with a person bending over nearby. &
\adjustbox{valign=c}{\includegraphics[height=40pt, width=50pt]{images/examples/idmr_cand.jpg}} & 1.0 & 0.1k\\

\midrule
WikiPerson~\cite{wikiperson} & \ding{52}&
\adjustbox{valign=c}{\includegraphics[height=40pt, width=40pt]{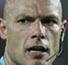}} &
Represent the face in the given image. &
\adjustbox{valign=c}{\includegraphics[height=40pt, width=50pt]{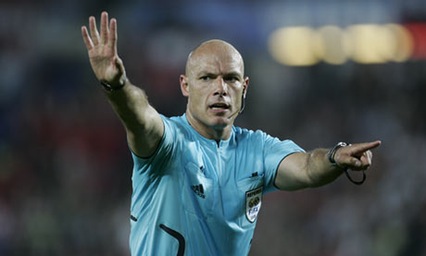}} & 1.0 & 3.0k  \\

\midrule
FORB~\cite{forb} & \ding{52} &
\adjustbox{valign=c}{\includegraphics[height=40pt, width=40pt]{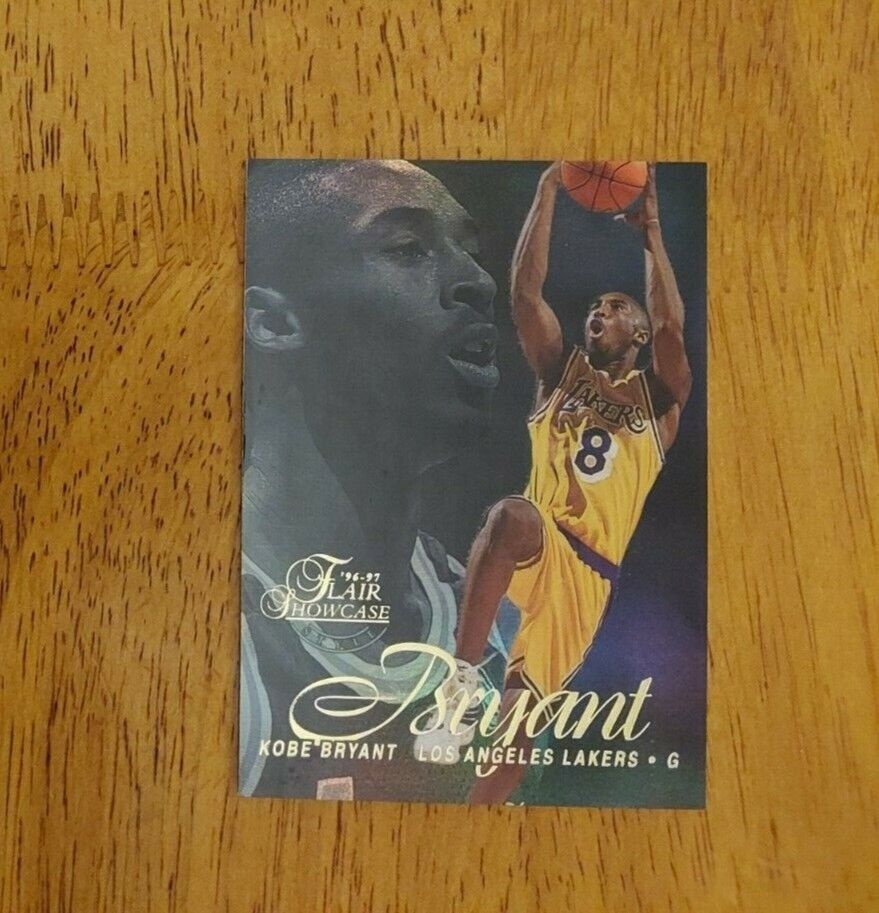}} &
Represent the image with the following text. Find the same object as the one in the given image. &
\adjustbox{valign=c}{\includegraphics[height=40pt, width=50pt]{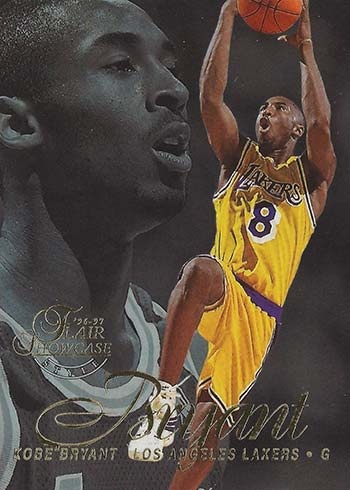}} & 1.0 & 20.0k \\

\midrule
IDMROOD~\cite{idmr} & \ding{52} &
\adjustbox{valign=c}{\includegraphics[height=40pt, width=40pt]{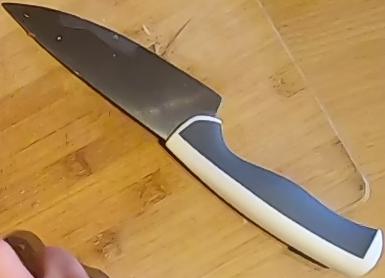}} &
Given the knife in the image, find an everyday image that contains the knife: &
\adjustbox{valign=c}{\includegraphics[height=40pt, width=50pt]{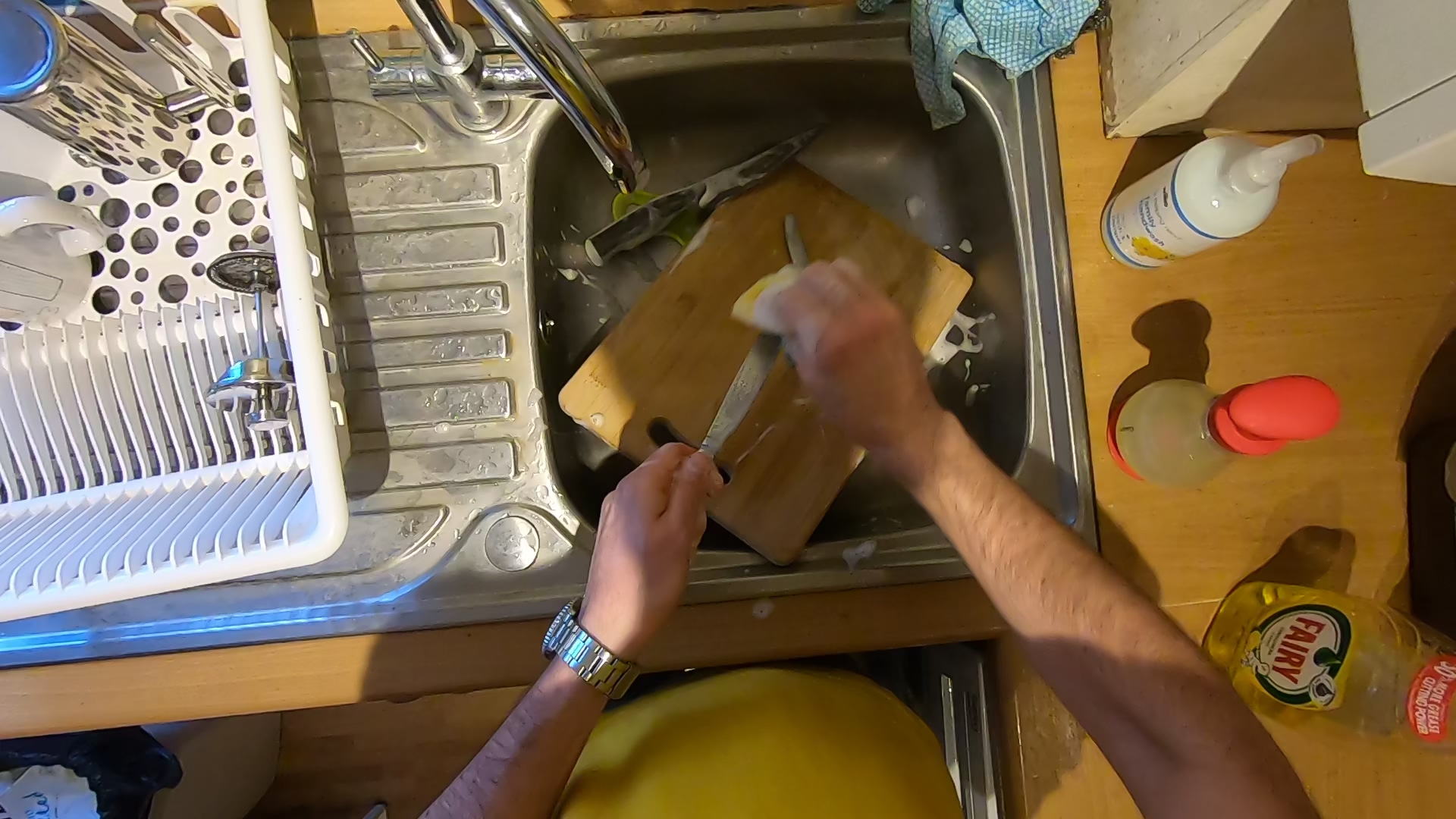}} & 1.0 & 1.0k\\

\bottomrule 
\end{tabularx}
\end{table*}

\begin{table*}[t] % 由于是table*，[t] 是最有效的选项
\centering
\caption{Examples of the identity editing task from MVEB.}
\label{tab:mveb_id_edit}
\begin{tabularx}{\linewidth}{@{}l c c X c@{} c c}
\toprule
\textbf{Dataset} & \textbf{OOD?} & \textbf{Query Image} & \textbf{Query Text} & \textbf{Target Image} & $\quad\boldsymbol{N_{tgt}}$ & $\boldsymbol{N_{cand}}$  \\
\midrule
GPTImageEdit~\cite{gptimageedit} & \ding{55} &
\adjustbox{valign=c}{\includegraphics[height=40pt, width=40pt]{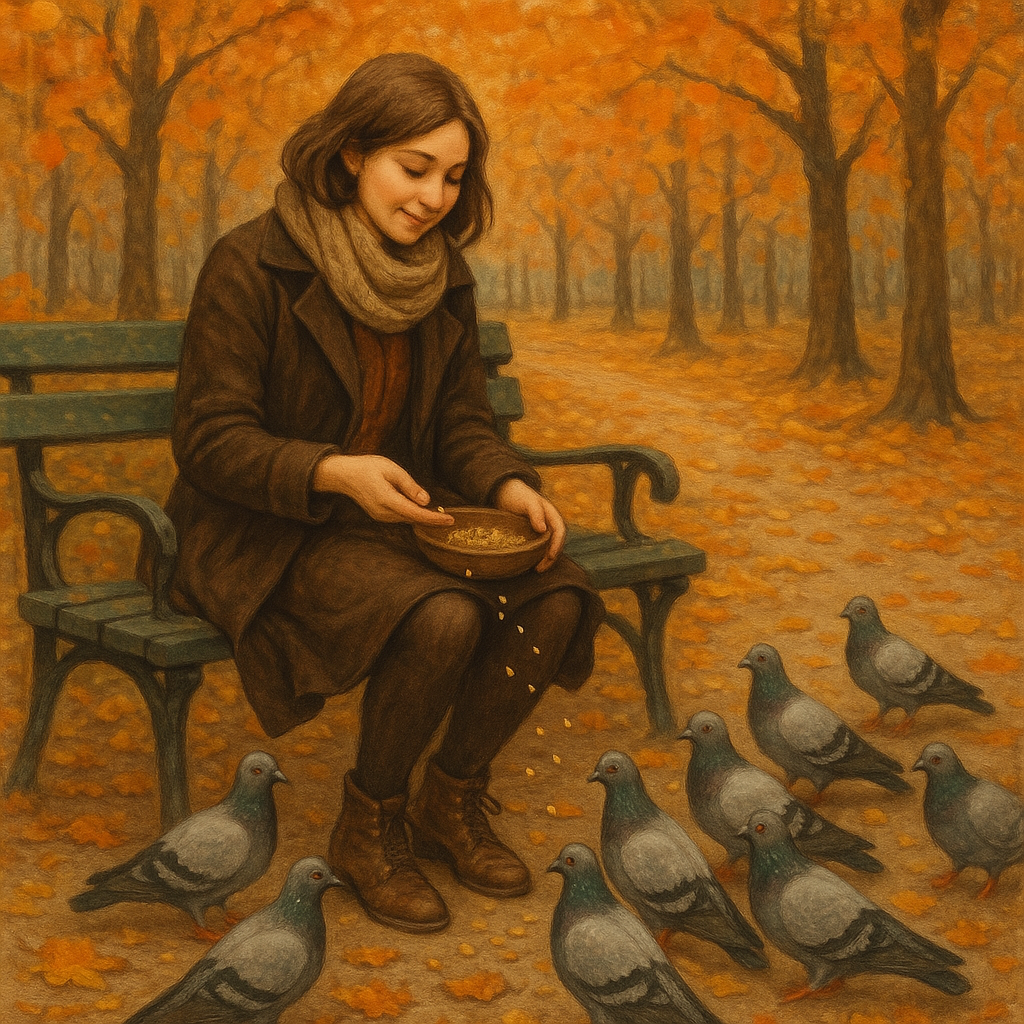}} &
Represent the image with the following editing text. Replace the pigeons with squirrels.&
\adjustbox{valign=c}{\includegraphics[height=40pt, width=40pt]{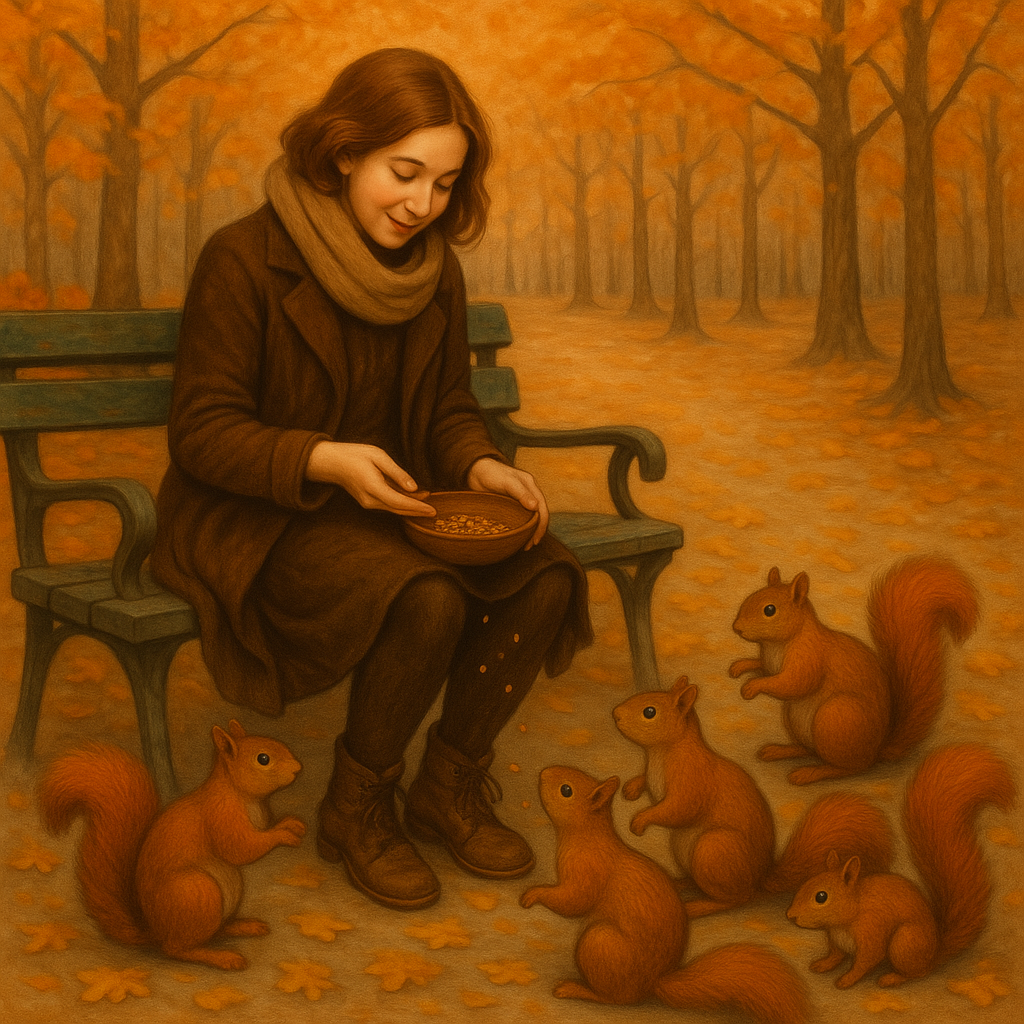}} & 1.0 & 2.9k  \\

\midrule
SEEDMultiTurn~\cite{seed} & \ding{55} &
\adjustbox{valign=c}{\includegraphics[height=40pt, width=40pt]{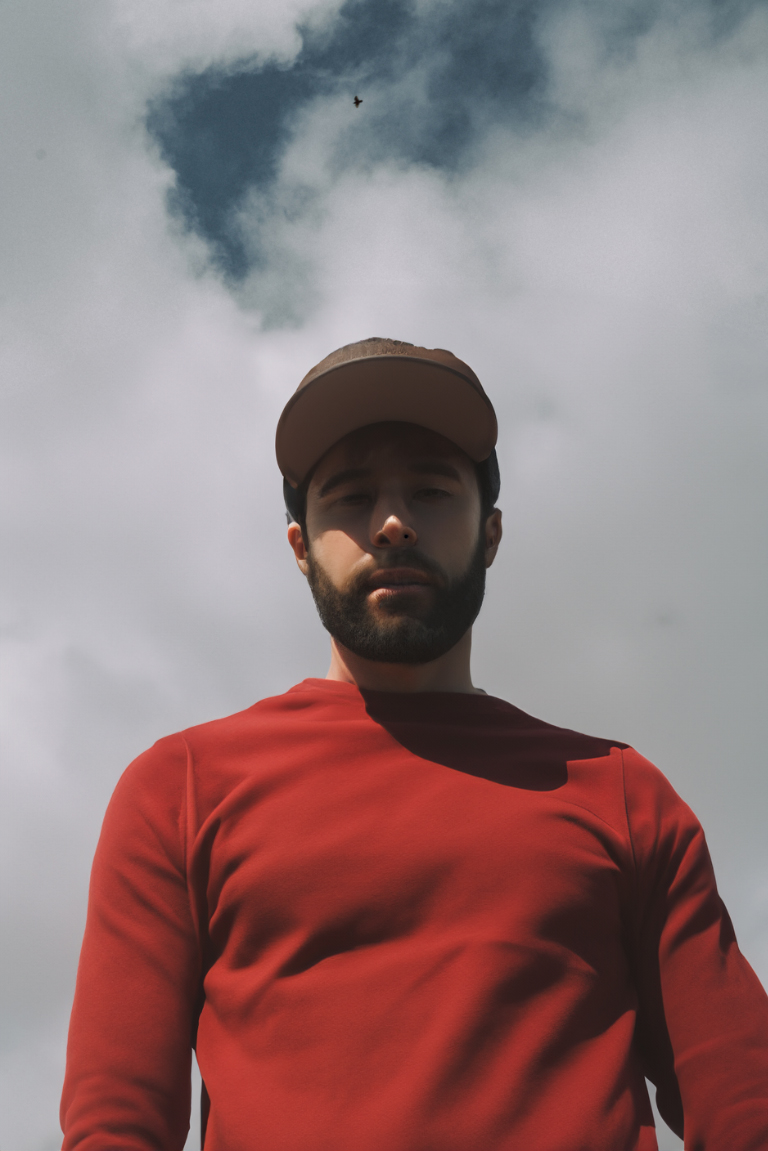}} &
Represent the image with the following editing text. Change the background to a street background.&
\adjustbox{valign=c}{\includegraphics[height=40pt, width=40pt]{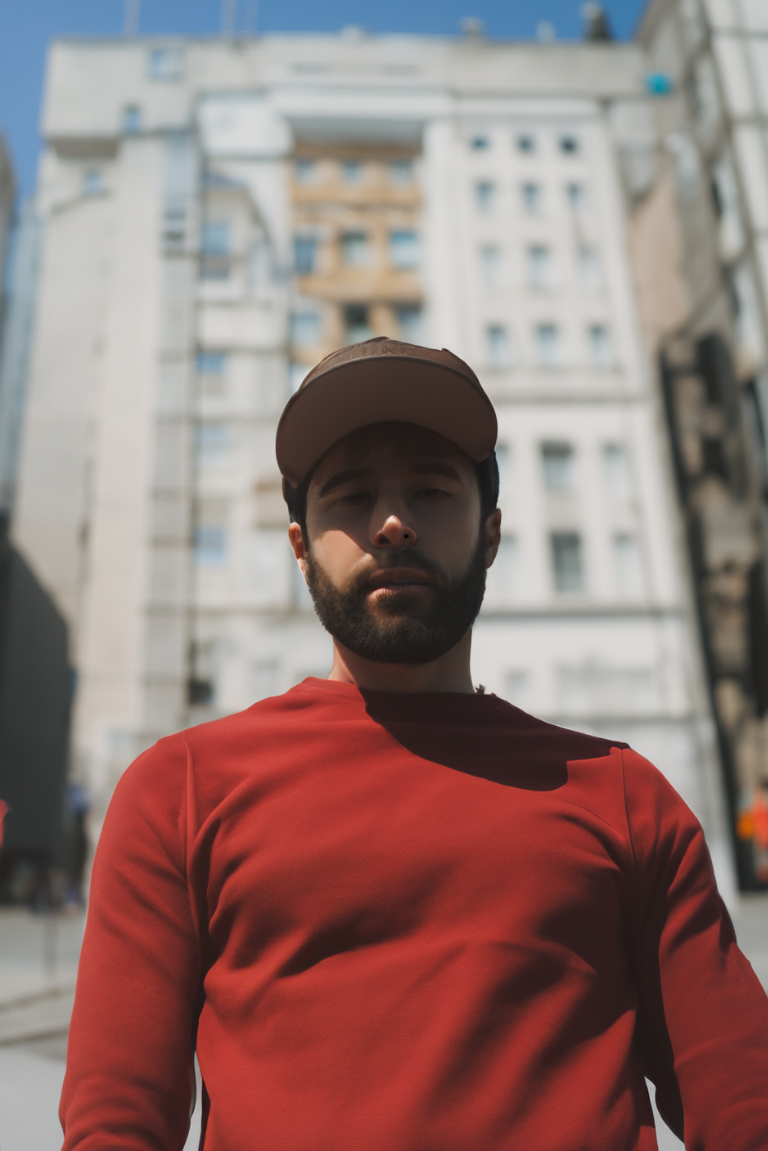}} & 1.0 & 3.1k \\

\midrule
COCOEdit~\cite{uniworld} & \ding{55} &
\adjustbox{valign=c}{\includegraphics[height=40pt, width=40pt]{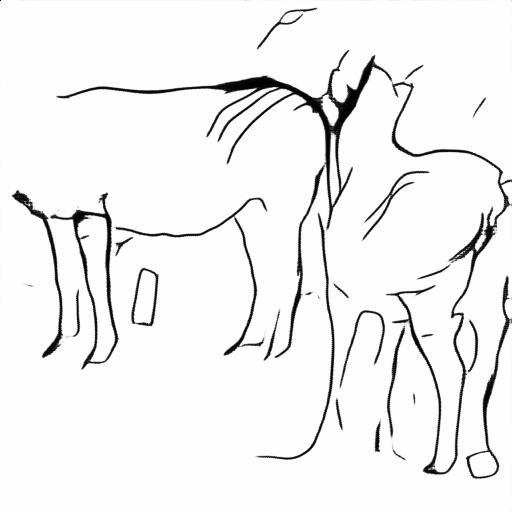}} &
Represent the image with the following editing text. There are two zebras standing in a field.. &
\adjustbox{valign=c}{\includegraphics[height=40pt, width=40pt]{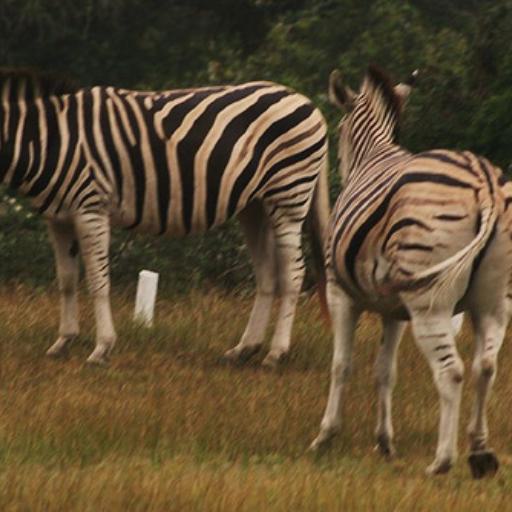}} & 1.0 & 2k \\

\midrule
SynCPR~\cite{syncpr} & \ding{55} &
\adjustbox{valign=c}{\includegraphics[height=40pt, width=40pt]{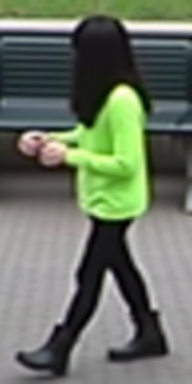}} &
Find the same person described by the following text. Wearing white sneakers, sitting on a park bench. &
\adjustbox{valign=c}{\includegraphics[height=40pt, width=40pt]{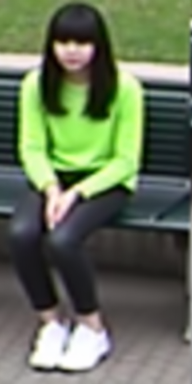}} & 1.0 & 3.0k\\

\midrule
OpenGPT4o~\cite{opengpt4o} & \ding{52} &
\adjustbox{valign=c}{\includegraphics[height=40pt, width=40pt]{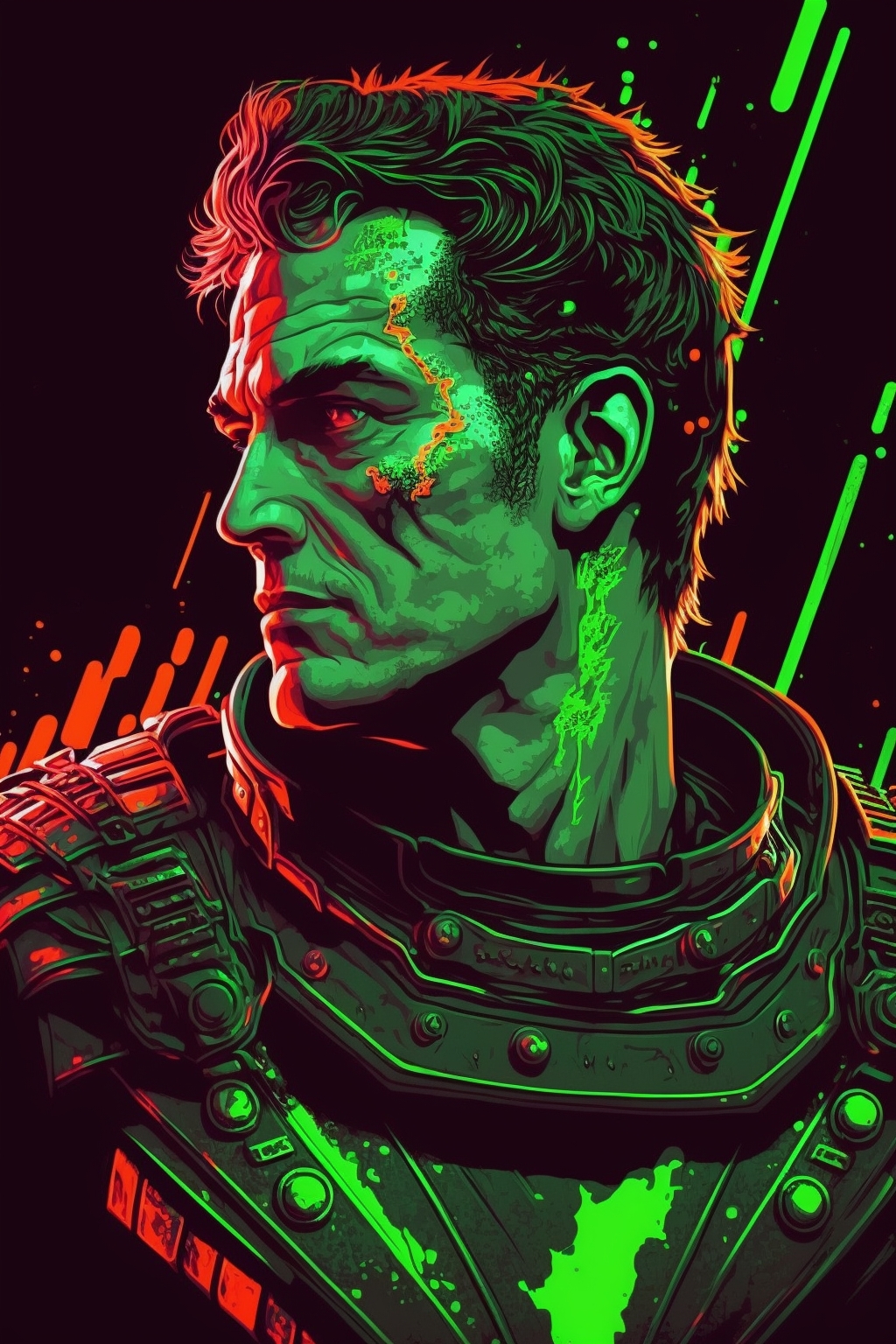}} &
Represent the image with the following text. Add a futuristic visor to the character's eyes. &
\adjustbox{valign=c}{\includegraphics[height=40pt, width=40pt]{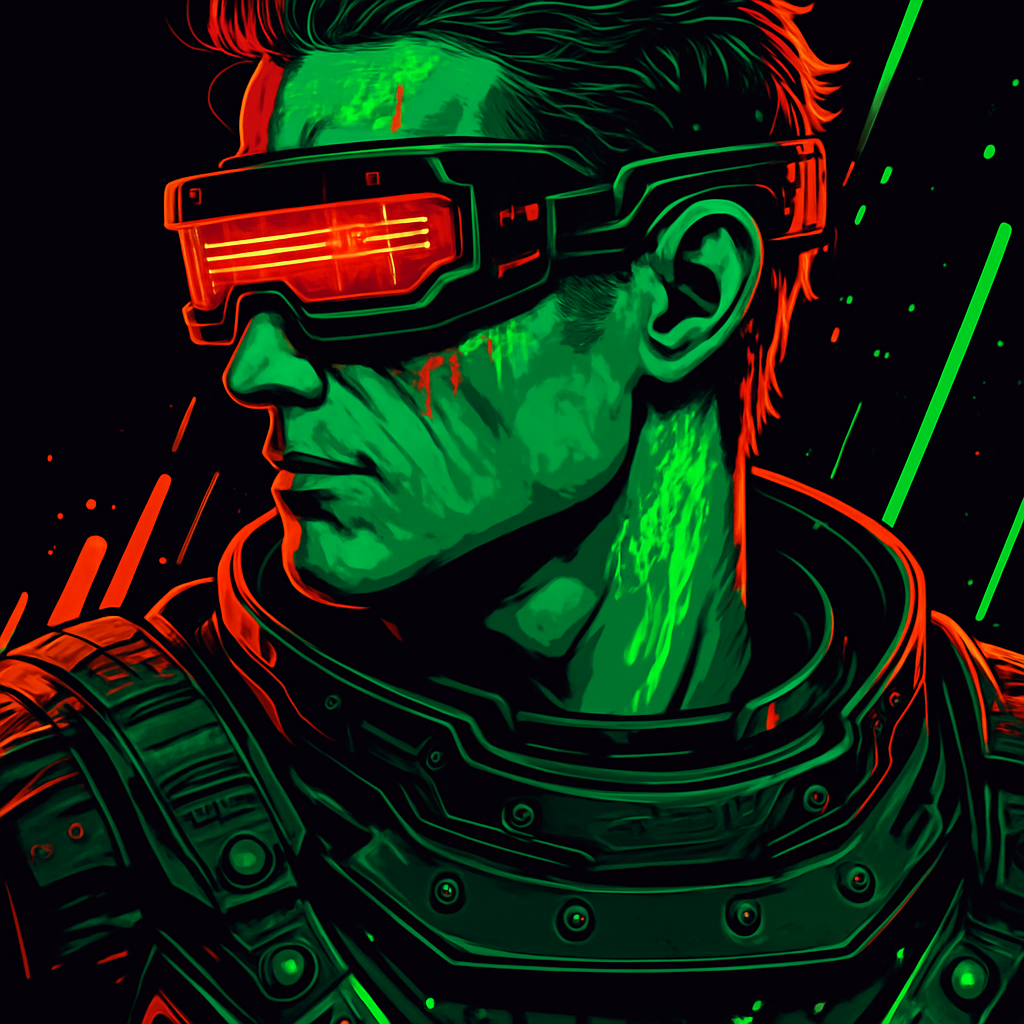}} & 1.0 & 20.0k \\

\bottomrule 
\end{tabularx}
\end{table*}

\section{Examples of MVEB Dataset}
In this section, we present one representative sample from each sub-dataset of our MVEB dataset, as summarized in Tables~\ref{tab:mveb_id_rec}, \ref{tab:mveb_reid}, \ref{tab:mveb_id_grd}, \ref{tab:mveb_id_edit}. The columns in these tables are defined as follows:
\begin{itemize}
    \item \textbf{OOD?}: Indicates whether the dataset is out-of-distribution.
    \item \textbf{Query Image}: The input query image.
    \item \textbf{Query Text}: The input query text.
    \item \textbf{Target Image}: The ground-truth target image corresponding to the query.
    \item $\boldsymbol{N_{tgt}}$: The average number of target items per query within the candidate set.
    \item $\boldsymbol{N_{cand}}$: The total size of the candidate pool.
\end{itemize}

These representative examples illustrate the diversity and structure of the MVEB dataset, providing a clear reference for understanding the characteristics and scale of each sub-dataset.
\clearpage 